 \documentclass[10pt,journal,compsoc]{./IEEEtran}

\usepackage{booktabs}
\usepackage{color}
\usepackage{xcolor}
\usepackage{times}
\usepackage{epsfig}
\usepackage{graphicx}
\usepackage{graphics}
\usepackage{amsmath}
\usepackage{amssymb}
\usepackage{multirow}
\usepackage{mathrsfs}
\usepackage{rotating}
\usepackage{bbm}
\usepackage{verbatim}
\usepackage{array}
\usepackage{booktabs}
\usepackage{tabularx}
\usepackage{url}
\usepackage{marvosym}
\usepackage[hidelinks]{hyperref} 
\hypersetup{hypertex=true,
            colorlinks=true,
            linkcolor=red,
            anchorcolor=blue!58,
            citecolor=blue!58
            }
\usepackage{CJKutf8}

\definecolor{mygray}{RGB}{218, 218, 218}
\definecolor{diff}{RGB}{0, 0, 0}
\usepackage{wrapfig}
\usepackage{colortbl}
\PassOptionsToPackage{table}{xcolor}
\usepackage[table]{xcolor}

\newcommand{\todo}[1]{\textcolor{red}{{[TODO: #1]}}}

\def\eg{\emph{e.g}.}

\def\etal{\emph{et al}.}
\def\ie{\emph{i.e}.}

\ifCLASSOPTIONcompsoc
  \usepackage[nocompress]{cite}
\else
  \usepackage{cite}
\fi
\ifCLASSINFOpdf
\else
\fi
\begin{document}
\title{
Spatial Frequency Modulation for \\ Semantic Segmentation
}
\author{Linwei~Chen,
		Ying~Fu,~\IEEEmembership{Senior Member,~IEEE},
		Lin~Gu,
        Dezhi~Zheng,
        Jifeng~Dai
\IEEEcompsocitemizethanks{
\IEEEcompsocthanksitem Linwei~Chen and Ying~Fu are with MIIT Key Laboratory of Complex-field Intelligent Sensing, Beijing Institute of Technology, Beijing, China, and School of Computer Science and Technology, Beijing Institute of Technology, Beijing, China.
\IEEEcompsocthanksitem Lin~Gu is with RIKEN AIP, Tokyo, Japan, and RCAST, The University of Tokyo, Tokyo, Japan.
\IEEEcompsocthanksitem Dezhi~Zheng is with MIIT Key Laboratory of Complex-field Intelligent Sensing, Beijing Institute of Technology, Beijing, China, and Instrumentation and Optoelectronic Engineering, Beihang University.
\IEEEcompsocthanksitem Jifeng~Dai is with Department of Electronic Engineering, Tsinghua University, Beijing, China.
}
\thanks{

$\textrm{\Letter}$ Ying Fu: fuying@bit.edu.cn
}
}


%
%

\markboth{IEEE Transactions on Pattern Analysis and Machine Intelligence}%
{Shell \MakeLowercase{\textit{et al.}}: Bare Demo of IEEEtran.cls for Computer Society Journals}
%



\IEEEtitleabstractindextext{%
\begin{abstract}
High spatial frequency information, including fine details like textures, significantly contributes to the accuracy of semantic segmentation. 
However, according to the Nyquist-Shannon Sampling Theorem, high-frequency components are vulnerable to aliasing or distortion when propagating through downsampling layers such as strided-convolution. Here, we propose a novel Spatial Frequency Modulation (SFM) that modulates high-frequency features to a lower frequency before downsampling and then demodulates them back during upsampling.
Specifically, we implement modulation through adaptive resampling (ARS) and design a lightweight add-on that can densely sample the high-frequency areas to scale up the signal, thereby lowering its frequency in accordance with the Frequency Scaling Property.
We also propose Multi-Scale Adaptive Upsampling (MSAU) to demodulate the modulated feature and recover high-frequency information through non-uniform upsampling
This module further improves segmentation by explicitly exploiting information interaction between densely and sparsely resampled areas at multiple scales. 
Both modules can seamlessly integrate with various architectures, extending from convolutional neural networks to transformers. 
Feature visualization and analysis demonstrate that our method effectively alleviates aliasing while successfully retaining details after demodulation. As a result, the proposed approach considerably enhances existing state-of-the-art segmentation models (e.g., Mask2Former-Swin-T +1.5 mIoU, InternImage-T +1.4 mIoU on ADE20K). 
Furthermore, ARS also enhances the performance of powerful Deformable Convolution (+0.8 mIoU on Cityscapes) by maintaining relative positional order during non-uniform sampling. 
Finally, we validate the broad applicability and effectiveness of SFM by extending it to image classification, adversarial robustness, instance segmentation, and panoptic segmentation tasks.
The code is available at \href{https://github.com/Linwei-Chen/SFM}{https://github.com/Linwei-Chen/SFM}.
\end{abstract}

}

\maketitle

\IEEEdisplaynontitleabstractindextext

%
\IEEEpeerreviewmaketitle


%
%
%
%

 

\IEEEraisesectionheading{\section{Introduction}\label{sec:introduction}}
\label{sec:intro}
\IEEEPARstart{A}{s} one of the fundamental tasks in computer vision, semantic segmentation plays an important role in various applications, including autonomous driving~\cite{2023planning}, robotics~\cite{2019bonnet}, and urban planning~\cite{2020gid}.
It aims to densely assign a semantic category for each pixel~\cite{2016fcn, deeplabv3plus, 2021segformer}, relying on spatial high-frequency information that includes fine details such as textures, patterns, structures, and boundaries~\cite{2019pattern, 2021learningstructure, 2021statisticaltexture, 2019boundaryaware}.

{

\color{diff}
However, high-frequency components have long been suspected to be vulnerable in deep neural network (DNN) based segmentation models. Specifically, modern DNNs~\cite{resnet2016, 2017resnext, 2021swin, 2022convnext} usually incorporate a series of downsampling operations after convolution or transformer blocks. While effective for expanding the receptive field and reducing dimensionality, these standard downsampling operations (such as pooling layers~\cite{vgg, resnet2016}, strided-convolutions~\cite{resnet2016, 2023internimage}) inherently risk information loss, particularly for high-frequency details. Critically, for frequencies exceeding the Nyquist limit, this process leads to aliasing. Aliasing occurs when a signal's high-frequency components are downsampled with an insufficient sampling rate, causing them to be misrepresented as lower frequencies. According to the Nyquist-Shannon Sampling Theorem \cite{shannon1949communication, nyquist1928}, frequencies higher than the Nyquist frequency (which is equivalent to half of the sampling rate) inherently lead to aliasing during uniform downsampling. For instance, frequencies above $\frac{1}{4}$ get aliased during a standard 2$\times$ downsampling operation (\eg, a 1$\times$1 convolution layer with a stride of 2 has a sampling rate of $\frac{1}{2}$)~\cite{2022flc}. Recent works~\cite{2020delving, 2022flc} have also empirically demonstrated a strong correlation between these high-frequency aliasing artifacts and the vulnerability of DNNs under adversarial attack~\cite{2016PGD, 2020autoattack}, highlighting a broader consequence of improper frequency handling.
}

\begin{figure}[tb!]
\centering
\scalebox{1.28}{
\includegraphics[width=0.68\linewidth]{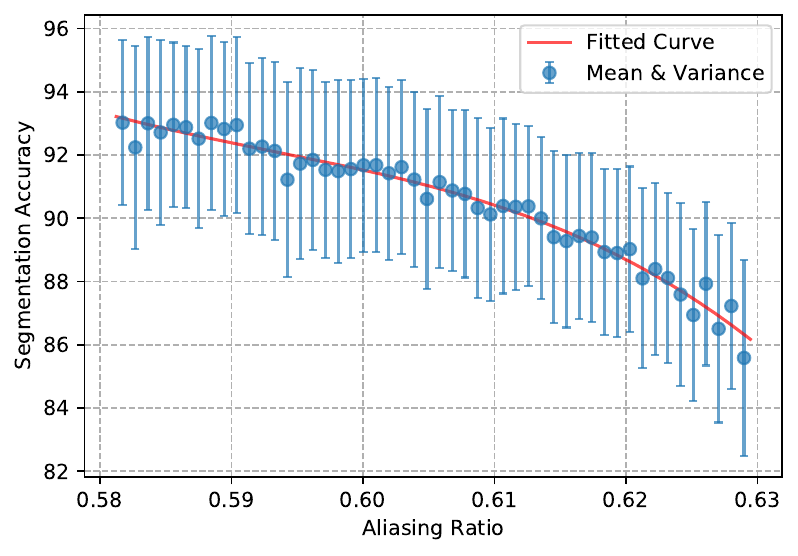} 
}  
\vspace{-3.98mm}  
\caption{
Quantitative analysis of the relationship between segmentation accuracy and the aliasing ratio.
Aliasing ratio is defined as the ratio of high frequency above the Nyquist frequency to the total power of the spectrum.  
We observe that image samples with a lower aliasing ratio in the features tend to exhibit better segmentation accuracy.
This phenomenon occurs in widely used existing models, including ResNet-50~\cite{resnet2016} (here), Swin-Transformer~\cite{2021swin}, and ConvNeXt~\cite{2022convnet} (see Figure~\ref{fig:HFPR3Models}).
}
\label{fig:freq_acc}
\vspace{-3.98mm} 
\end{figure}

\begin{figure}[tb!]
\centering
\includegraphics[width=0.98\linewidth]{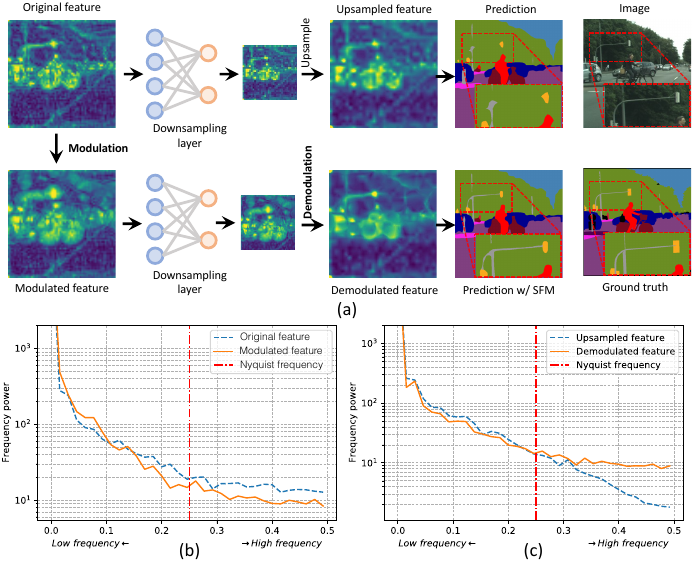}
\vspace{-3.98mm} 
\caption{
(a) Illustration of Spatial Frequency Modulation.
(b) 
The modulated feature contains fewer high frequencies above the Nyquist frequency, thus mitigating aliasing.
(c) 
High frequencies tend to decay significantly in normally upsampled features, whereas the demodulated feature successfully recovers more high-frequency details.
}
\label{fig:SFM}
\vspace{-5.18mm} 
\end{figure}

{\color{diff}

To delve deeper into the connection between frequency handling and segmentation performance, beyond the known \textit{spectral bias} of DNNs~\cite{2019spectralbias, 2021deepfrequencyprinciple}, we quantitatively investigate the specific impact of aliasing. Using Fourier analysis, we propose the ``aliasing ratio" metric – defined as the ratio of spectral power above the Nyquist frequency to the total spectral power within a feature map. This metric directly quantifies the proportion of high-frequency energy susceptible to aliasing during subsequent downsampling. A higher aliasing ratio signifies a greater potential for distortion. Our analysis in Figure~\ref{fig:freq_acc} reveals a crucial finding: segmentation accuracy consistently decreases as the aliasing ratio increases across various architectures. We term this phenomenon ``\textit{aliasing degradation}," highlighting the detrimental effect of excessive high-frequency content relative to the sampling rate. This empirical evidence strongly motivates the need to manage the frequency distribution in feature maps before downsampling to mitigate performance loss.
}

{\color{diff}
In signal processing, applying low-pass filters to the input signal before sampling is one general solution to avoid aliasing~\cite{1987digital}. 
Interestingly, this coincides with the success of a group of methods~\cite{2019makingshiftinvariant, 2020delving, 2023depthadablur, 2022flc} which apply a low-pass filter either in the spatial domain~\cite{2019makingshiftinvariant, 2020delving, 2023depthadablur} or in the spectral domain~\cite{2022flc} to the feature map.
Although effective in relieving aliasing, directly removing high-frequency content sacrifices fine details and textures which are crucial for accurate scene parsing and understanding.
}

{\color{diff}

Rather than discarding high frequencies via filtering, which sacrifices potentially valuable details, we propose an alternative approach: Spatial Frequency Modulation (SFM). SFM introduces a novel two-stage process specifically designed to combat aliasing degradation while preserving information. First, a \textit{modulation} operation adaptively resamples the feature map before downsampling, effectively shifting high-frequency components to lower frequencies, as depicted in Figure~\ref{fig:SFM} (a) and (b). This pre-emptive frequency shift largely reduces the aliasing ratio, allowing more spectral energy to pass through the subsequent downsampling layer without distortion. Second, after the feature has traversed the network's downsampling path, a \textit{demodulation} operation during upsampling aims to reverse the initial modulation, recovering the high-frequency details that were temporarily shifted, as shown in Figure~\ref{fig:SFM} (a) and (c). This modulate-demodulate cycle is the core concept of SFM, offering a way to protect high-frequency information from aliasing instead of simply removing it. Consequently, by better preserving fine details like boundaries. SFM leads to improved segmentation accuracy.
}

\begin{figure}[tb!]
\centering
\includegraphics[width=0.98\linewidth]{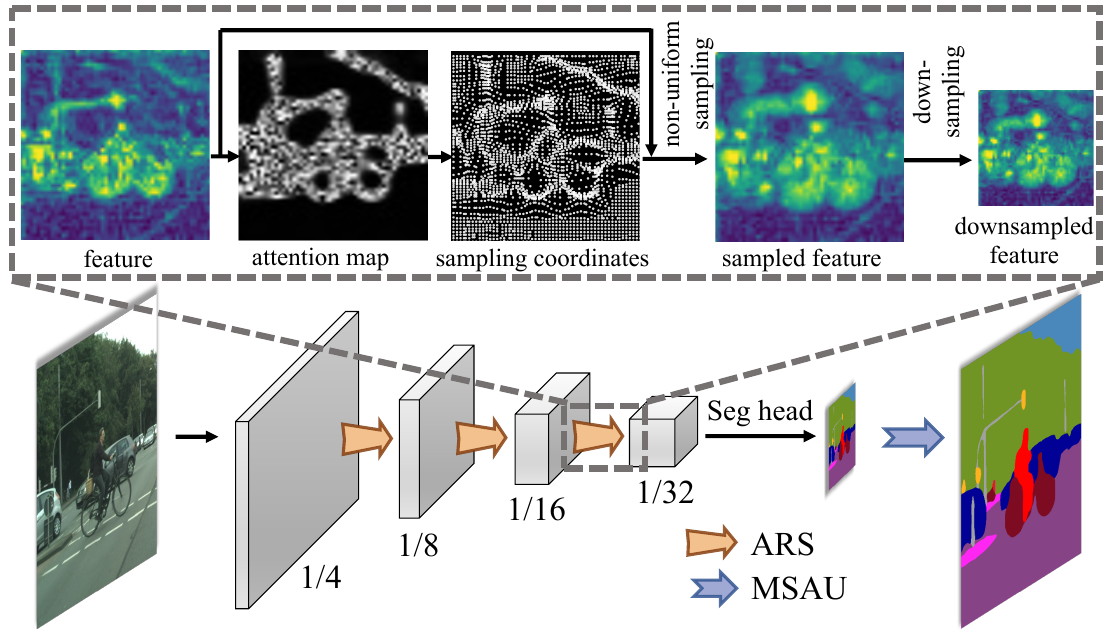} 
\vspace{-3.98mm}     
\caption{
Illustration of a general FCN-based architecture for semantic segmentation.
The proposed Adaptive Resampling (ARS) and Multi-Scale Adaptive Upsampling (MSAU) can be seamlessly integrated into existing models.
}
\label{fig:overview}
\vspace{-3.98mm} 
\end{figure}

Since aliasing occurs during downsampling, our modulation adopts an adaptive sampling strategy~\cite{2022learningtodown, 2023lzu}, which is a lightweight, out-of-the-box add-on operation that can easily be applied to CNNs and Transformers.
As shown in Figure~\ref{fig:overview}, modulation consists of the following steps:
1) Given a feature map, a high-frequency-sensitive attention generator highlights areas with high-frequency information through high attention values.
2) Sampling coordinates are then assigned adaptively based on the attention map, resulting in denser sampling on areas with high attention values.
3) The feature map is subsequently resampled using these coordinates, generating a non-uniformly sampled feature map of the same size, which we refer to as the modulated feature.
Specifically, according to the Frequency Scaling Property, densely sampling the signal with a sampling rate of $A$ results in its frequency decreasing to $\frac{1}{A}$~\cite{1987digital}.

{\color{diff}
Segmentation models~\cite{2016fcn, pspnet, 2022pcaa, 2022mask2former} typically employ a progressive downsampling of feature maps to generate a smaller prediction map, which is post-upssampled to restore the original resolution.
In our approach, we replace the standard post-upsampling with MSAU. This involves demodulating the low-resolution modulated feature/prediction using MSAU, bringing it back to the original size.
It consists of two steps: 
1) Non-uniform upsampling. Traditional semantic segmentation networks~\cite{2016fcn, pspnet, 2022pcaa} often produce a small score map on downsampled features before upsampling to the original input size. They typically achieve this by uniformly upsampling the prediction map through bilinear interpolation. However, our adaptive resampling modulates feature frequency by deforming spatial coordinates, which necessitates the preservation of pixel alignment during demodulation. 
Non-uniform upsampling achieves this through triangulation and barycentric interpolation. Specifically, for each pixel in the upsampled result, its value on the uniform grid is barycentrically interpolated from the nearest three pixels in the non-uniformly represented (modulated) features. 
2) Multi-scale relation mining. 
This module dynamically builds the relation between local pixels in the densely and sparsely sampled area and refines the prediction by incorporating multi-scale local information.
}

{\color{diff}
\begin{itemize}
\item 
We propose Spatial Frequency Modulation (SFM), a framework consisting of modulation and demodulation operations. Modulation operation modulates high frequency in feature to lower frequency before downsampling, alleviating aliasing. 
Demodulation operation reverses the modulated feature and recovers more high-frequency details.
\item We implement the SFM framework using adaptive resampling (ARS) and multi-scale adaptive upsampling (MSAU). These two lightweight modules can be easily incorporated into existing models and optimized with the standard end-to-end training procedure. 
\item Extensive experiments verify that our method consistently improves various state-of-the-art semantic segmentation architectures, ranging from CNN-based InternImage to Transformer-based Swin-Transformer, with only a minor additional computational burden. 
Furthermore, its general applicability has been confirmed in image classification, robustness against adversarial attacks, as well as instance/panoptic segmentation.
\end{itemize}

Our main contributions can be summarized as follows:
\begin{itemize}
    \item { We identify and quantify the ``aliasing degradation" phenomenon in semantic segmentation. To mitigate this degradation, we propose Spatial Frequency Modulation (SFM), a novel conceptual framework to mitigate aliasing degradation. SFM involves \textit{modulating} high frequencies to lower bands before downsampling and \textit{demodulating} them back during upsampling, aiming to preserve rather than discard high-frequency details.}
    \item { We introduce Adaptive Resampling (ARS) for modulation and Multi-Scale Adaptive Upsampling (MSAU) for demodulation, providing a practical and lightweight implementation of the SFM framework. These modules seamlessly integrate into existing CNN and Transformer architectures with minimal overhead and standard end-to-end training.}
    \item { We demonstrate through extensive experiments that SFM consistently enhances various state-of-the-art segmentation models (e.g., +1.5 mIoU on Mask2Former-Swin-T, +1.4 mIoU on InternImage-T for ADE20K). We further validate its effectiveness in improving model robustness and its generalizability to image classification, instance, and panoptic segmentation.}
\end{itemize}
}

\section{Related Work}
%

\noindent{\bf Semantic segmentation.}
With the development of Deep Learning~\cite{2020canet, 2021ctnet, 2020refinenet, 2023context, 2018deep, zhang2024deep, tian2023transformer, fu2015reflectance, wei2021physics, fuying-2021-tpami, 2022Guided, 2022gan, 2022levelAware, chen2022consistency, 2023casid, liu2024transformer}, semantic segmentation has long been dominated by FCN-based models and their transformer-based successors~\cite{2016fcn, chen2023semantic, 2022hybridsupervised, 2021efficienthybrid, 2021maskformer, 2022mask2former}.
To address the issue of high-frequency detail loss resulting from downsampling, mainstream state-of-the-art methods adopt two different strategies, \ie, retaining high-resolution feature maps and fusing low-level feature maps.
HRNet~\cite{hrnet} and HRFormer~\cite{hrformer} incorporate multiple branches at different scales, including a high-resolution branch designed to preserve spatial details.
Dilated residual networks (DRN)~\cite{2017dilated} remove downsampling layers and rely on dilated convolutions to maintain a large receptive field. 
Building upon the foundation of dilated residual networks, PSPNet~\cite{pspnet} introduces the pyramid pooling module (PPM) to model multi-scale contexts. 
In contrast, the DeepLab series~\cite{deeplabv3, deeplabv3plus} employ atrous spatial pyramid pooling (ASPP). 
Non-local networks ~\cite{2018nonlocalnetworks}, CCNet ~\cite{2020ccnet}, and DANet ~\cite{danet} adopt a self-attention mechanism to capture pixel-wise context within a global perceptual field.
PCAA~\cite{2022pcaa} simultaneously utilizes local and global class-level representations for attention calculation.
However, such methods often incur high computational costs to retain high-resolution feature maps.

Alternatively, FPN-based models~\cite{2017feature, kirillov2019panopticfpn, 2019unetnature, 2020pointrend} leverage the lateral path to fuse detailed information from high-resolution feature maps at the low level.
SFNet~\cite{2020semanticflow}, AlignSeg ~\cite{2021alignseg}, and FaPN ~\cite{2021fapn} further enhance the FPN structure by aligning features at different scales.
Different from existing work, the proposed spatial frequency modulation framework directly improves the downsampling operation by optimizing the frequency distribution.
It can also be easily integrated into both existing DRN-based semantic segmentation methods~\cite{2016fcn, pspnet, 2020ccnet, 2021ocnet, 2022pcaa} and FPN-based methods~\cite{2021fapn, 2022mask2former}. 

\vspace{+0.518mm}
\noindent{\bf Attention mechanism.}
Currently, extensive works exploit attention mechanisms~\cite{2017attention} for segmentation. DANet~\cite{danet} proposes a dual attention module to establish a global connection between pixels spatially and across channels. 
CCNet~\cite{2020ccnet} and OCNet~\cite{2021ocnet} improve the efficiency of spatial attention. SETR~\cite{2021setr}, Segmentor~\cite{2021segmenter}, and Segformer~\cite{2021segformer} integrate the self-attention of transformers into the semantic segmentation architecture to achieve a global receptive field.
Different from these works, which enhance deep features numerically by highlighting the importance of feature values, the proposed method enhances deep features spatially by enlarging high-frequency feature areas.



\vspace{+0.518mm}
\noindent{\bf Anti-Aliasing.}
Recently, researchers have explored various methods to mitigate aliasing effects in DNNs. 
Some approaches, such as~\cite{2019makingshiftinvariant}, focus on enhancing shift-invariance by employing anti-aliasing filters implemented as convolutions. 
Other methods, like~\cite{2020delving}, take a step further by incorporating learned blurring filters to improve shift-invariance, and \cite{2023depthadablur} adopts a depth adaptive blurring filter and incorporates an anti-aliasing activation function.
Additionally, \cite{2021wavecnet} leverages the low-frequency components of wavelets to reduce aliasing and enhance robustness against common image corruptions. 
Beyond image classification, anti-aliasing techniques have gained relevance in the domain of image generation. 
In \cite{2021aliasgan}, blurring filters are utilized to remove aliases during image generation in generative adversarial networks (GANs), while~\cite{2020watchupconv} and~\cite{2021spectral} employ additional loss terms in the frequency space to address aliasing issues. 
Motivated by these findings, this paper further observes an interesting phenomenon called the ``aliasing degradation" and proposes a new concept called spatial frequency modulation to address it. 
Different from these works, the proposed method modulates high frequency to lower frequency instead of directly removing them, benefiting the subsequent  recovery of the high frequency details.

{
\vspace{+0.518mm}
\noindent{\bf Adaptive sampling.}
Uniform sampling has long been considered a standard operation in computer vision. However, this approach has been suspected to be inefficient due to the fact that not all image regions are equally important, and important regions are usually sparse~\cite{2020spatially, 2020dyconv}. 
Recently, non-uniform sampling has been proposed to sample more informative regions in images~\cite{2015stn, 2018learningtozoom, 2019efficientseg, 2019devilindetials, 2022learningtodown}.
Saliency sampler~\cite{2018learningtozoom, 2022learningtodown, 2021fovea} and trilinear attention sampling network~\cite{2019devilindetials} estimate the spatial saliency or attention map first and then ``zoom-in" the task-relevant image areas.
Besides, deformable convolutional networks~\cite{2017deformable, 2019deformablev2} improve standard convolution by making it deformable to the shape of object. 
SSBNet~\cite{2022ssbnet} adopts saliency sampler in bottlenecks of ResNet~\cite{resnet2016}.
AFF~\cite{2023AFF} abandons the classic grid structure and develops a point-based local attention block, facilitated by a balanced clustering module and a learnable neighborhood merging.

Compared to the most relevant existing work ES~\cite{2019efficientseg} and its improved versions~\cite{2022learningtodown, 2023lzu}, ES~\cite{2019efficientseg} is specifically designed for image downsampling and is applied to efficient low-resolution semantic segmentation. Therefore, it is {\it not valid} for full-resolution semantic segmentation. On the other hand, the proposed method can generally improve semantic segmentation and {is applicable to any resolution, including low-resolution scenarios}.
}

\vspace{+0.518mm}
\noindent{\bf Frequency learning.}
Nowadays, many researchers are shifting their focus to frequency learning~\cite{2024freqfusion, 2025fdconv, 2024fadc, chen2025frequency}. 
Rahaman~\etal~\cite{2019spectralbias} find that these networks prioritize learning the low-frequency modes, a phenomenon called the spectral bias. 
Xu~\etal~\cite{2021deepfrequencyprinciple} conclude the deep frequency principle, the effective target function for a deeper hidden layer biases towards lower frequency during the training~\cite{2021deepfrequencyprinciple}.
Qin~\etal~\cite{2021fcanet} exploring utilizing more frequency for channel attention mechanism.
Xu~\etal~\cite{2020learningfrequency} introduce learning-based frequency selection method into well-known neural networks, taking JPEG encoding coefficients as input.
Huang~\etal~\cite{2023adaptivefrequency} employ the conventional convolution theorem in DNN, demonstrating that adaptive frequency filters can efficiently serve as global token mixers.
Compared to existing works, this work investigates the relationship between high frequency and segmentation accuracy for the first time by using aliasing ratio, demonstrating the possibility of improving segmentation accuracy by optimizing the frequency distribution of features.

{
\section{Method}
In this section, we first present our quantitative analysis of the relationship between feature frequency and segmentation using the aliasing ratio. We also discuss how this analysis motivates our research. Next, we delve into the specifics of Adaptive Resampling (ARS) and Multi-Scale Adaptive Upsampling (MSAU), the techniques we employ to realize the Spatial Frequency Modulation (SFM) framework. Finally, we analyze the frequency distribution in modulated and demodulated features to validate the effectiveness of the proposed method.
}

\begin{figure}[tb!]
\centering
\scalebox{0.98}{
\begin{tabular}{ccccc}
\includegraphics[width=0.958\linewidth]{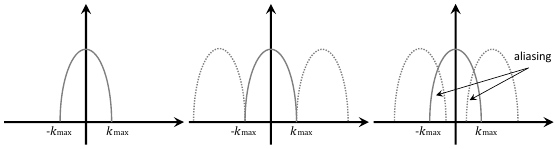} 
\vspace{-3.98mm} 
\end{tabular}
}   
\caption{
Illustration of 1-D signal aliasing in the frequency domain.
Left: The frequency spectrum of a 1D signal with a maximal frequency $k_{\text{max}}$ in the continuous domain.
Center: After sampling, replicas of the signal appear at distances proportional to the sampling rate. If the signal is sampled with a sufficiently large sampling rate (\eg, $>$ Nyquist rate), the signal in the discrete domain does not overlap with its replicas.
Right: If the signal is undersampled, its replicas overlap and lead to aliasing.
}
\label{fig:aliasing_1d}
\vspace{-4mm} 
\end{figure}

\begin{figure}[tb!]
\centering
\scalebox{0.918}{
\begin{tabular}{ccccc}
\includegraphics[width=0.5\linewidth]{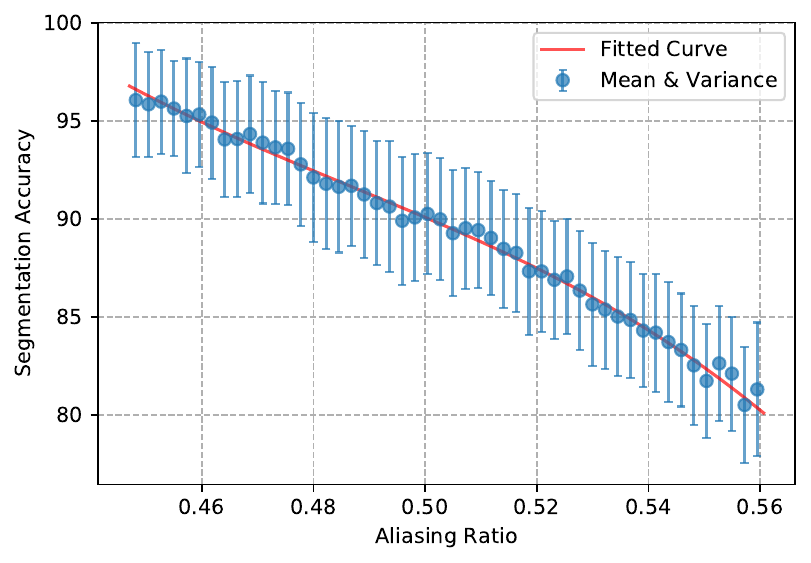} 
&\includegraphics[width=0.5\linewidth]{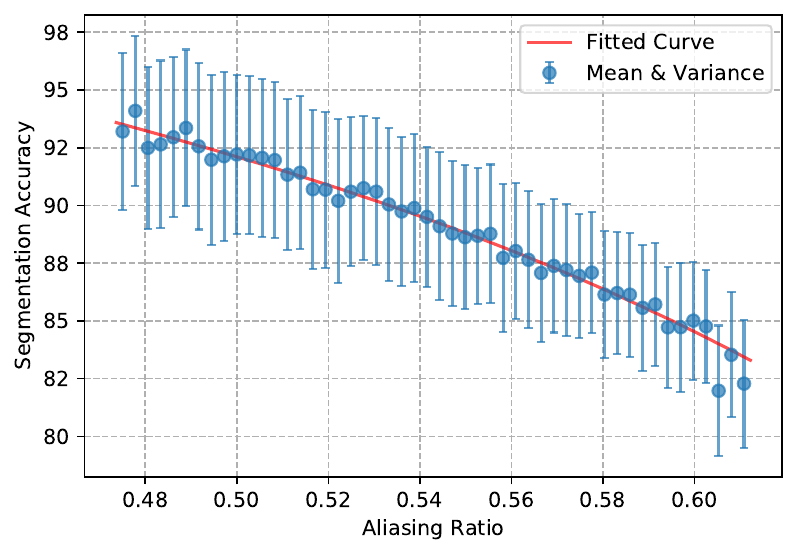} 
\\
\vspace{-2.98mm} 
(a) Swin-Transformer
&(b) ConvNeXt
\end{tabular}
}   
\caption{
We conduct an aliasing ratio analysis for existing models including ResNet~\cite{resnet2016}, Swin Transformer~\cite{2021swin}, and ConvNeXt~\cite{2022convnet}.
}
\label{fig:HFPR3Models}
\vspace{-3.98mm} 
\end{figure}

\begin{figure}[t!]
\centering
\includegraphics[width=0.98\linewidth]{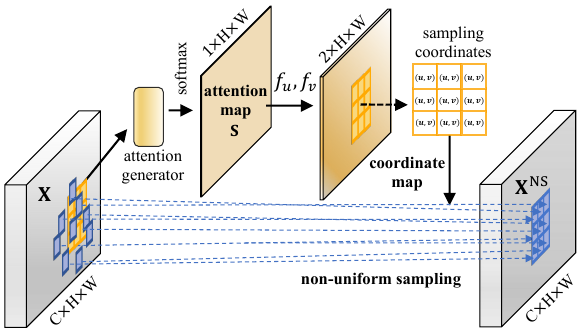}
\vspace{-3.98mm}    
\caption{
Illustration of adaptive resampling (ARS).
We illustrate how it resamples 9 points adaptively for brevity.
}
\label{fig:sampler}
\vspace{-3.98mm} 
\end{figure}

\begin{figure}[t!]
\centering
\begin{tabular}{c}
\includegraphics[width=0.98\linewidth]{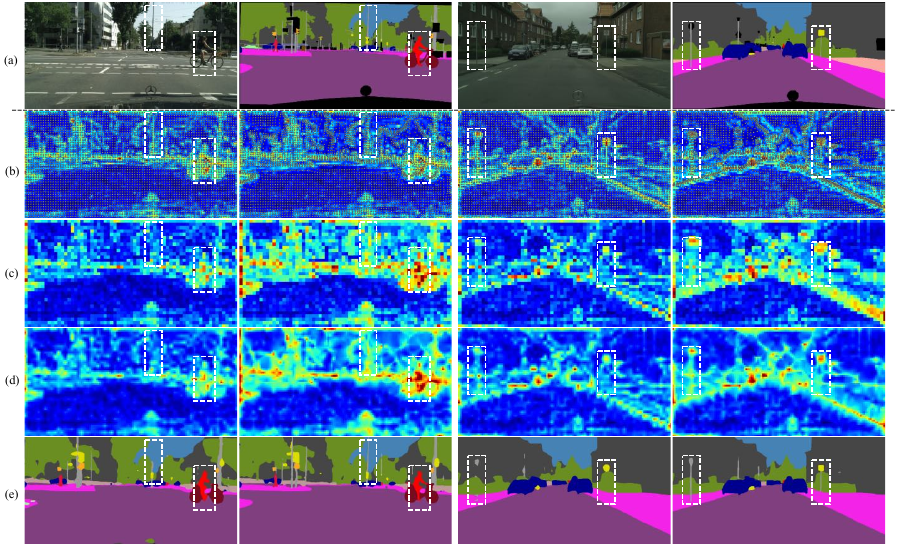} \\
\footnotesize
\hspace{+1mm} w/o SFM \hspace{+8mm}\quad w/ SFM \hspace{+8mm}\quad w/o SFM \hspace{+8mm}\quad w/ SFM
\vspace{-2.98mm} 
\end{tabular}
\caption{
Feature visualization.
(a) shows the input image and ground truth.
(b) visualizes the original feature map. 
White points indicate sampling coordinates. 
Our method samples the object boundaries more densely.
(c) and (d) shows the downsampled and recovered feature map.
(e) are final predictions.
Zoom in for a better view.
}
\label{fig:featurevis2}
\vspace{-3.98mm} 
\end{figure}

{
\subsection{Analysis with Aliasing Ratio}
Regardless of their specific network topology, modern DNNs essentially consist of a sequence of stacked convolution or transformer blocks and downsampling layers~\cite{resnet2016, 2021swin, 2022convnet}. The downsampling is commonly achieved through strided-convolutions or pooling layers.

From a signal processing perspective, the inherent spatial pyramid design of most DNNs evidently violates the fundamental signal processing principle known as the Sampling Theorem~\cite{shannon1949communication, nyquist1928}, particularly in their downsampling operations~\cite{2019makingshiftinvariant, 2020delving, 2022flc}. 
Specifically, the majority of architectures downsample feature maps without ensuring that the sampling rate exceeds the Nyquist rate~\cite{shannon1949communication, nyquist1928}. Consequently, this leads to a phenomenon referred to as aliasing, where high frequencies above the Nyquist frequency during downsampling overlap in the frequency spectra. These overlaps introduce ambiguities where high-frequency components cannot be distinctly differentiated from low-frequency components, as illustrated in Figure~\ref{fig:aliasing_1d}.
Although this paradigm is suspected to have underlying issues, clear evidence to prove its harmfulness for standard segmentation has not yet been found.
} 

To investigate the intricate relationship between high frequency and segmentation accuracy, we introduce the metric termed aliasing ratio to quantitatively measure the level of high frequency. 
The aliasing ratio is defined as the ratio of high frequency power above the Nyquist frequency to the total power of the spectrum. 
Specifically, to obtain the aliasing ratio, we first need to transform the feature map to the frequency domain using Discrete Fourier Transform (DFT). 
After performing a DFT on the feature map $f\in \mathbb{R}^{M\times N}$, it can be represented as:
\begin{equation}
\begin{aligned}
F(k, l) = \frac{1}{MN}\sum_{m=0}^{M-1}\sum_{n=0}^{N-1}f(m, n)e^{-2\pi j (\frac{k}{M}m + \frac{l}{N}n)}.
\end{aligned}
\end{equation}
{\color{diff}
Here, $F \in \mathbb{C}^{M \times N}$ represents the output complex-valued array from the DFT, where $M$ and $N$ denote the height and width of the input feature map $f$, respectively. The indices $m$ and $n$ represent the spatial coordinates within the feature map $f(m, n)$.
}
The frequencies in the height and width dimensions are given by $\left|\frac{k}{M}\right|$ and $\left|\frac{l}{N}\right|$, with $k$ taking values from the set $\{0, 1, \ldots, M-1\}$ and $l$ from $\{0, 1, \ldots, N-1\}$.
According to the Sampling Theorem~\cite{shannon1949communication}, frequencies above the Nyquist frequency, which is half of the sampling rate ($\frac{sr}{2}$), result in aliasing.
In many existing modern models~\cite{resnet2016, 2021swin, 2022convnet}, a common downsampling rate of 2 is used, resulting in a sampling rate of $\frac{1}{2}$ and a Nyquist frequency of $\frac{1}{4}$.
Thus, the set of high frequencies larger than the Nyquist frequency is $\mathcal{H} = \{(k, l) \mid |\frac{k}{M}| > \frac{1}{4} \text{ or } |\frac{l}{N}| > \frac{1}{4}\}$.
Consequently, the aliasing ratio can be formulated as follows:
\vspace{-1mm}
\begin{equation}
\begin{aligned}
\label{eq:aliasing_ratio}
\text{Aliasing Ratio} &= \frac{\sum_{(k, l)\in \mathcal{H}}\left|F_{}(k, l) \right|}{\sum_{}\left|F_{}(k, l) \right|}.
\end{aligned}
\vspace{-1mm}
\end{equation}
In practice, we shift the low-frequency components into the center of the array, resulting in $F_s$. 
The aliasing ratio is quantitatively calculated by subtracting the low-frequency power from the center of the array, denoted as $F_s[\frac{M}{4}:\frac{3M}{4}, \frac{N}{4}:\frac{3N}{4}]$ from the total power and then calculating the ratio.

{
Using this metric, we perform an analysis on widely used models such as ResNet~\cite{resnet2016}, as well as recent state-of-the-art transformer-based and convolution-based models like Swin Transformer~\cite{2021swin} and ConvNeXt~\cite{2022convnet} on the high-resolution Cityscapes dataset~\cite{cityscapes2016}.
As depicted in Figure~\ref{fig:HFPR3Models}, the mean segmentation accuracy gradually decreases as the aliasing ratio increases.

This observation reveals an inverse correlation between the ratio of high frequencies above the Nyquist frequency and segmentation accuracy. This suggests that we can enhance existing segmentation models by adjusting the frequency distribution in the feature to be downsampled.
Motivated by this insight, we introduce the Spatial Frequency Modulation (SFM) framework, comprising two key operations: modulation and demodulation. The proposed SFM offers the following advantages:
1) During the modulation operation, by modulating the high frequencies in the feature map to be downsampled to a lower frequency, the aliasing ratio is significantly reduced, thus mitigating aliasing. This not only enhances segmentation accuracy but also improves robustness against adversarial attacks.
2) During the demodulation operation, more high frequencies can be recovered from the modulated feature, encouraging the model to learn more discriminative features containing higher frequencies. This leads to better segmentation details.
We implement the modulation and demodulation operations using adaptive resampling (ARS) and multi-scale adaptive upsampling (MSAU), respectively. Next, we provide details of their implementation."
}



\subsection{Adaptive Resampling}
\label{sec:sampler}

We implement the modulation operation using Adaptive Resampling (ARS), as illustrated in Figure~\ref{fig:sampler}. 
Adaptive sampling strategies have previously been employed for efficient image downsampling~\cite{2022learningtodown, 2023lzu}, which is beneficial for efficient inference.
Here, we utilize ARS to densely resample the high-frequency regions within the feature map. This reduces the rate of pixel value fluctuations in the corresponding areas of the original feature map, effectively modulating the high frequency to a lower frequency. 
Specifically, on the basis of Frequency Scaling Property, scaling a signal up by a factor of $A$ can be achieved by densely sampling the signal with a sampling rate of $A$, resulting in its frequency decreasing to $\frac{1}{A}$~\cite{1987digital,2007firstcourse}.
Thus, for a specific frequency of $F(k, l)$, aliasing can be avoided during 2$\times$ downsampling if we choose the scaling factors/sampling rate $(A_k, A_l)$ to make:
\begin{equation}
\begin{aligned}
\label{eq:scaling}
\frac{1}{A_k} \times \left| \frac{k}{M}\right| \leq \frac{1}{4}, \quad \frac{1}{A_l} \times \left|\frac{l}{N}\right| \leq \frac{1}{4}.
\end{aligned}
\end{equation}

{\color{diff}
Even though this method can mathematically ensure the avoidance of aliasing, it still gives rise to following problems:

1) If we densely sample/upsample a feature map with a sampling rate of $A_k>1$ and $A_l>1$ in a uniform manner, the height and width of the feature map increase to $(A_k, A_l)$ times the original size, resulting in increased computational costs for subsequent layers in the network.

To address this issue, the proposed ARS method adaptively samples the feature map without altering its size. Specifically, it sparsely samples low-frequency areas, such as smooth roads and sky regions, while conserving sampling coordinates for dense sampling of high-frequency areas, such as object boundaries.

2) Under the constraint of not altering the size of the feature map, it is difficult to calculate the precise adaptive sampling coordinates to ensure Equation~\eqref{eq:scaling}, due to the spatially varying signal frequency within the feature map.
}

{
To simplify the intricate calculation of adaptive sampling coordinates, we choose to directly supervise the adaptively resampled features. The objective is to reduce high frequencies above the Nyquist frequency, ultimately resulting in a lower aliasing ratio. During training, the ARS module becomes adept at predicting optimal sampling coordinates, effectively minimizing high frequencies in the output modulated feature. As shown in Figure~\ref{fig:featurevis2}(c), adaptive resampled features demonstrate a smoother appearance and retain more useful responses to the object of interest.
}

{\color{diff}
To guide the learning process of the ARS module, we introduce two distinct loss functions, \ie, the Frequency Modulation (FM) loss and the Semantic High Frequency (SHF) loss, both of which play a pivotal role.
The FM loss encourages the modulated feature to have fewer high frequencies in its frequency spectrum, prompting the adaptive resampler to shift high-frequency signals toward lower frequencies. This modulation applies to all high frequencies within the feature.
In contrast, the SHF loss specifically targets the high frequencies that are essential for the segmentation task, \ie, semantic boundaries. 
It guides the ARS module to densely sample the semantic high-frequency areas, effectively preserving object boundary information.
Specifically, as shown in Figure~\ref{fig:sampler}, the ARS consists of three steps:
1) Attention map generation. We use a lightweight attention generator to predict attention map $\mathbf{S}$ conditions on the input feature map $\mathbf{X}$.
2) Coordinates mapping. This step aims to map the uniform coordinate $(i, j)$ to the non-uniform sampling coordinate $(u, v)$ conditions on the attention map $\mathbf{S}$.  
3) Non-uniform sampling. The feature map is non-uniformly sampled with coordinates $(u, v)$, where areas with high attention values are sampled densely.
In the following, they are described in detail.
}

\vspace{+0.518mm}
\noindent{\bf Attention map generation.}
Given an input feature map $\mathbf{X}$ to be downsampled, we need to generate an attention map where high values correspond to high frequency areas.
To achieve this, we utilize difference-aware convolution (DAConv) to identify areas with significant pixel value variations and assign them a higher attention value.
The DAConv can be formulated as a central-to-surrounding difference operation:
\begin{equation}
\begin{aligned}
	\mathbf{X}'_{i, j} &= \sum_{p,q\in \Omega}
	\mathbf{W}_{p,q}^{} \cdot (\mathbf{X}_{i, j} - \mathbb{I}_{(p, q)\neq (0, 0)} \cdot \mathbf{X}_{i+p, j+q}), \\
\end{aligned}
\end{equation}
where $\mathbf{X}\in \mathbb{R}^{C\times H\times W}$, $\mathbf{W}_{p,q}$ is the weight of DAConv, $\mathbb{I}_{(p, q)\neq (0, 0)}$ is the indicator function that outputs 1 if $(p, q)\neq (0, 0)$, and 0 otherwise, and $\Omega$ represents offset set of the surrounding pixels.
$C$, $H$, and $W$ indicate the channel number, height, and width, respectively.
To make the output feature more stable and facilitate training, softmax normalization is applied kernel-wisely to ensure that the sum of the learned weights is constrained to 1: 
\begin{equation}
\begin{aligned}
	\mathbf{W}_{p,q} &= \frac{\exp(\mathbf{W}_{p,q})}{\sum_{p,q\in \Omega} \exp(\mathbf{W}_{p,q})}.
\end{aligned}
\end{equation}
Due to the central-surrounding difference calculation, DAConv is sensitive to local pixel changes, especially in high-frequency areas where neighboring pixel values often exhibit variations.

To capture multi-scale information and expand the perceptual field of the attention generator, we incorporate a lightweight Pyramid Spatial Pooling (PSP) module~\cite{pspnet}. The PSP module comprises four pyramid levels with bin sizes of 1, 2, 3, and 7 for each level. Following this, a convolutional layer is used. Finally, we perform a softmax operation along both height and width dimensions to obtain the attention map $\mathbf{S}\in \mathbb{R}^{H\times W}$.
This process can be formulated as: 
\vspace{-1mm}
\begin{equation}
\begin{aligned}
	\mathbf{S} &= \text{Softmax}\bigg(\text{Conv}\Big(\text{Concat}\big[\text{DAConv}(\mathbf{X}), \text{PSP}(\mathbf{X})\big]\Big). \\
\end{aligned}
\vspace{-1mm}
\end{equation}

{
\vspace{+0.518mm}
\noindent{\bf Coordinate mapping.}
We map the uniform coordinate $(i, j)$  to coordinate $(u, v)$ for adaptive sampling based on attention map $\mathbf{S}$
\begin{equation}
\begin{aligned}
u = f_{u}(i,j, \mathbf{S}),
v = f_{v}(i,j, \mathbf{S}),
\end{aligned}
\end{equation}
where $f_{u}$, $f_{v}$ are coordinate mapping functions.
{\color{diff}
Inspired by~\cite{2018learningtozoom}, the mapping functions effectively compute the new coordinates $(u, v)$ for a target location $(i, j)$ as a weighted average of neighboring source coordinates $(i', j')$. The weights are determined by the attention map values $\mathbf{S}(i', j')$ at those neighboring locations, modulated by a distance kernel $G((i, j), (i', j'))$. This mechanism naturally pulls the sampling grid towards regions with higher attention values, resulting in denser sampling in those areas. The distance kernel $G$ enforces a local constraint, ensuring that the influence of attention values diminishes with distance. The process is formulated as:
}
\begin{equation}
\begin{aligned}
\label{eq:coordinates_mapping}
\vspace{-1mm}
	f_u(i, j, \mathbf{S}) &= \frac{\sum_{i', j'}\mathbf{S}(i', j')G((i, j), (i', j'))i'}{\sum_{i', j'}\mathbf{S}(i', j')G((i, j), (i', j'))} ,\\
	f_v(i, j, \mathbf{S}) &= \frac{\sum_{i', j'}\mathbf{S}(i', j')G((i, j), (i', j'))j'}{\sum_{i', j'}\mathbf{S}(i', j')G((i, j), (i', j'))} ,\\
\end{aligned}
\vspace{-1mm}
\end{equation}
where $(i, j)\in[0,1]^2$, $(u, v)\in[0,1]^2$ are original coordinates and the non-uniform sampling coordinates.
Distance kernel $G$ is a fixed Gaussian function with a given standard deviation $\sigma+1$ and square shape of size $2\sigma+1$:
\begin{equation}
\begin{aligned}
\label{eq:gaussian}
G((i, j), (i', j')) = \sigma\sqrt{2\pi} \exp\left(-\frac{{(i' - i)^2 + (j' - j)^2}}{{2\sigma^2}}\right).
\end{aligned}
\end{equation}

This fixed Gaussian filter allows the calculated non-uniform sampling coordinates in Equation~\eqref{eq:coordinates_mapping} to be easily added to a standard network that preserves the differentiability needed for training.
Besides, we set
\begin{equation}
\begin{aligned}
\label{eq:coveringconstraint}
f_u(0, j, \mathbf{S}) = 0,\ &f_u(1, j, \mathbf{S}) = 1,  \\
f_v(i, 0, \mathbf{S}) = 0,\ &f_v(i, 1, \mathbf{S}) = 1.\\
\end{aligned}
\end{equation}
These constraints ensure the sampling coordinates cover the entire image, benefiting subsequent adaptive upsampling.
}

\vspace{+0.518mm}
\noindent{\bf Non-uniform sampling.}
With sampling coordinates $(u,v)$, the adaptive sampled feature can be obtained by
\begin{equation}
\begin{aligned}
&\mathbf{X}^{\text{NS}}(i, j) = \mathbf{X}(u, v),
\end{aligned}
\end{equation}
where $\mathbf{X}^{\text{NS}}(i, j)$ indicates non-uniform sampled feature at coordinates$(i, j)$, and $\mathbf{X}^{\text{NS}}\in \mathbb{R}^{C\times H\times W}$.
Since coordinates $(u, v)$ may be fractional, we follow~\cite{MaskRCNN2017} to use bilinear interpolation for sampling. 
In the modulated feature $\mathbf{X}^{\text{NS}}$, the high-frequency area is scaled up owing to denser sampling, effectively modulating it to a lower frequency. 

{
\vspace{+0.518mm}
\noindent{\bf Frequency modulation loss.}
We expect the ARS module to modulate high frequencies to lower frequencies, resulting in a reduction of the power of high frequencies in the frequency spectra of the modulated feature. Thus, we can regulate the module to minimize the frequency components above the Nyquist frequency during sampling:

\begin{equation}
\begin{aligned}
L_{FM} = \frac{1}{\left|\mathcal{H}\right|}\sum_{(k, l) \in \mathcal{H}}\left|F_{}(k, l) \right|^2,
\end{aligned}
\end{equation}
where $\mathcal{H} = \{(k, l) \mid |\frac{k}{M}| > \frac{1}{4} \text{ or } |\frac{l}{N}| > \frac{1}{4}\}$ is the set of high frequencies above the Nyquist frequency.
}

{
\vspace{+0.518mm}
\noindent{\bf Semantic high-frequency loss.}
To further enhance semantic boundary information during downsampling, we introduce the semantic high-frequency loss $L_{\text{SHF}}$ to supervise the ARS. 
Initially, we extract high-frequency areas from the ground truth semantic segmentation label:
\vspace{-1mm}
\begin{equation}
\begin{aligned}
\mathbf{\hat S} = \text{Laplacian}(\mathbf{\hat Y}),
\end{aligned}
\vspace{-1mm}
\end{equation}
where $\mathbf{\hat Y}$ is the ground truth segmentation label.
We use $3\times 3$ high-pass Laplacian filter that extracts high-frequency areas in $\mathbf{\hat Y}$.
For clearer semantic boundaries, Gaussian blur with standard deviation $\sigma = 1$ is applied to $\mathbf{\hat Y}$. Using the coordinate mapping function in Equation~\eqref{eq:coordinates_mapping}, we could get sampling coordinates based on $\mathbf{\hat S}$, where semantic boundaries are densely sampled:
\begin{equation}
\begin{aligned}
\hat u = f_u(i, j, \mathbf{\hat S}), \hat v = f_v(i, j, \mathbf{\hat S}).
\end{aligned}
\end{equation}
{\color{diff}
The semantic high-frequency loss $L_{\text{SHF}}$ encourages ARS to densely sample the semantic high-frequency areas
\begin{equation}
\begin{aligned}
L_{\text{SHF}} = \Vert u - \hat u \Vert_{2} + \Vert v - \hat v \Vert_{2}. \\
\end{aligned}
\end{equation}
}
Finally, the total loss for end-to-end training becomes
\begin{equation}
\begin{aligned}
L_{\text{total}} = L_{\text{seg}} + \lambda_{\text{FM}} L_{\text{FM}} + \lambda_{\text{SHF}} L_{\text{SHF}},
\end{aligned}
\end{equation}
where $L_{\text{total}}$, $L_{\text{seg}}$, $\lambda_{\text{FM}}$ and $\lambda_{\text{SHF}}$ are the total loss, segmentation loss, and weights for loss balancing, respectively.
$L_{\text{seg}}$ is the original standard loss for semantic segmentation and is usually a pixel-wise cross-entropy loss~\cite{2016fcn, pspnet, 2022pcaa}.
}


\begin{figure}[t!]
\centering
\includegraphics[width=0.98\linewidth]{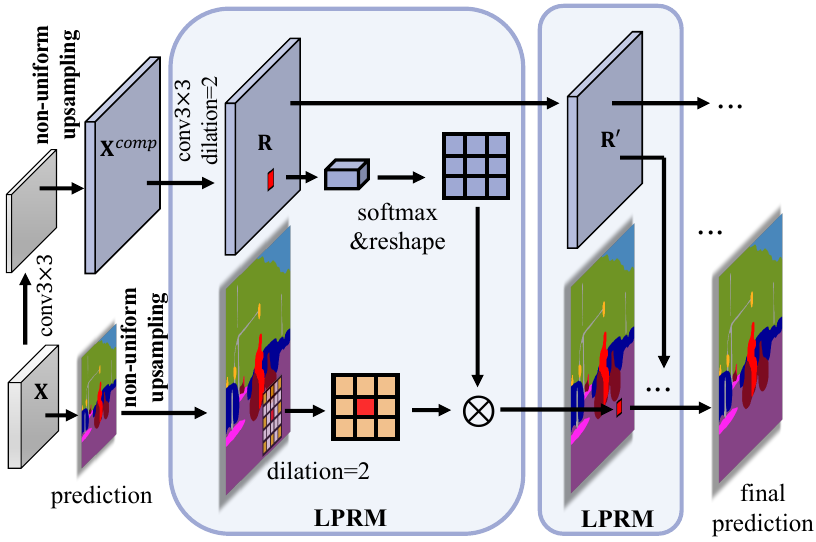}    
\vspace{-2mm}     
\caption{
Illustration of multi-scale adaptive upsampling. For brevity, we demonstrate two cascaded Local Pixel Relation Modules (LPRM) with a dilation of 2. Additional LPRMs can be similarly cascaded.
}
\label{fig:MSAU}
\vspace{-3.98mm} 
\end{figure}

\begin{figure}[t!]
\centering
\includegraphics[width=0.38\linewidth]{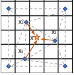}    
\vspace{-4mm}     
\caption{
Illustration of non-uniform upsampling.
The blue points represent known adaptively sampled points, and the red star indicates a point in the target non-uniformly upsampled feature. Its value is obtained through barycentric interpolation from the three nearest pixels in the non-uniformly represented (modulated) features.
}
\label{fig:tri_interp}
\vspace{-3.98mm}  
\end{figure}
\begin{figure}[tb!]
\centering
\scalebox{0.78}{
\begin{tabular}{ccccc}
\includegraphics[width=0.58\linewidth]{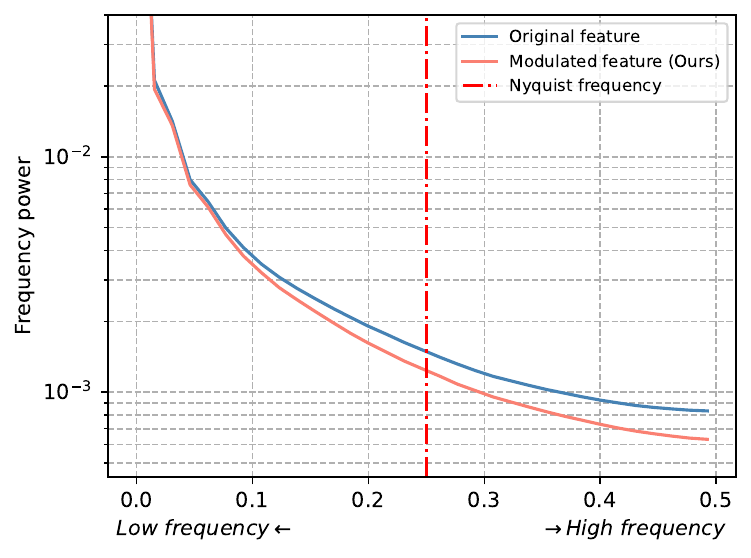}
& \includegraphics[width=0.58\linewidth]{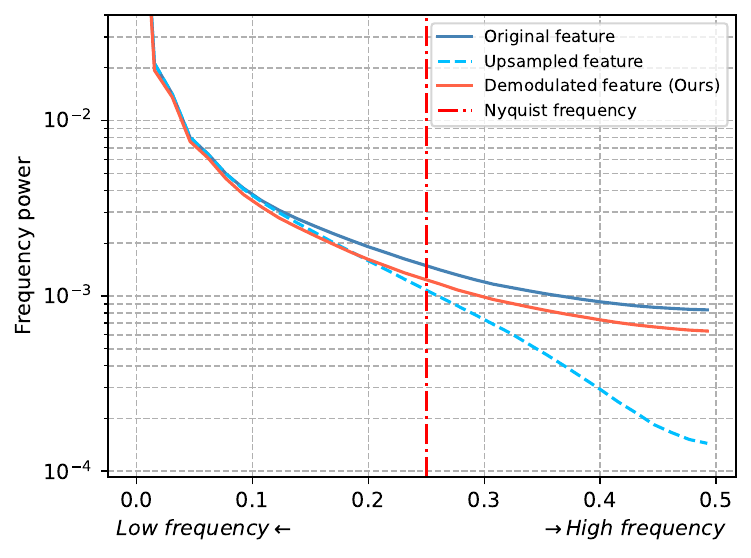}  
\\
(a) Modulated feature
&(b) Demodulated feature
\vspace{-2.98mm} 
\end{tabular}
}   
\caption{
{
Feature frequency analysis. 
(a) demonstrates a reduction in high frequencies for features modulated by ARS.
(b) The demodulated feature analysis shows that it retains more high frequencies.
}
}
\label{fig:frequency_analysis}
\vspace{-5.18mm} 
\end{figure}

\begin{figure}[tb!]
\centering
\scalebox{0.998}{
\begin{tabular}{ccccc}
\hspace{-3mm}
\includegraphics[width=0.998\linewidth]{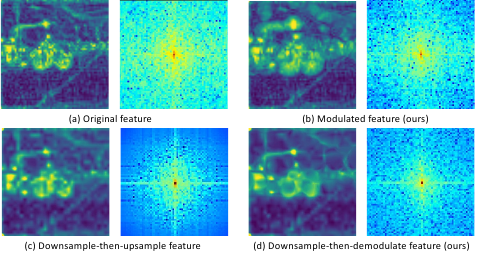} 
\vspace{-3.98mm} 
\end{tabular}
}   
\caption{
{
Feature visualization.
(a) displays the original feature map alongside its frequency spectra, where the center represents low frequency and the corners depict high frequency. Brighter colors indicate higher power of the corresponding frequency.
In (b), our modulated feature is presented, with densely sampled high-frequency areas from the original. Consequently, these areas are enlarged and appear smoother, resulting in darker high-frequency corners in the spectra.
After 2$\times$ downsampling, we upsample the downsampled feature to the original size in (c) and demodulate the modulated feature to the original size in (d). Due to downsampling, many high-frequency details are lost, resulting in a significant decay in the high-frequency area of the frequency spectra in (c).
Our demodulated feature in (d) retains more details, with the corners of the frequency spectra brighter than in (c), indicating a recovery of more high frequencies.
}
}
\label{fig:feature_freq_vis}
\vspace{-5.18mm} 
\end{figure}

\subsection{Multi-Scale Adaptive Upsampling}
{
In prior works~\cite{2016fcn, pspnet, 2022pcaa}, the segmentation head produces a segmentation map of the same size as its downsampled features.
To generate the final segmentation results, it is necessary to upsample the segmentation map to the input size.
Existing methods simply upsample downsampled predictions via uniform bilinear interpolation~\cite{pspnet, 2020ccnet, 2022pcaa}.
However, since our adaptive resampling deforms spatial coordinates, our strategy requires a non-uniform upsampling.
Moreover, in adaptive sampling, densely sampled areas usually contain richer information than sparsely sampled areas. 
We need to explicitly exploit information interaction between densely and sparsely sampled areas.
On the basis of the above analysis, 
we propose Multi-Scale Adaptive Upsampling (MSAU).
As shown in Figure~\ref{fig:MSAU}, MSAU firstly upsamples the prediction non-uniformly and then adaptively refines the prediction by utilizing multi-scale relation with a series of Local Pixel Relation Modules (LPRM). 
As shown in Figure~\ref{fig:featurevis2}(d), the demodulated features demonstrate more useful detailed responses to the object of interest, indicating the retention of valuable high-frequency information.
}

{\color{diff}
\vspace{+0.518mm}
\noindent{\bf Non-uniform upsampling.}
Non-uniform upsampling recovers an upsampled feature in a uniform grid from the modulated feature that is non-uniformly represented.
We employ Delaunay triangulation~\cite{1934delaunay} and barycentric interpolation within triangles~\cite{shirley2009fundamentals,1996quickhull}. Specifically, we first divide the known points at non-uniform sampling coordinates into multiple triangles and then perform linear interpolation inside them to obtain points at uniform coordinates.
The constraints in Equation~\eqref{eq:coveringconstraint} ensure that the convex hull of the sampling locations covers the entire image, thus allowing interpolation to recover results to the original input resolution.

As shown in Figure~\ref{fig:tri_interp}, for each pixel in the non-uniform upsampled feature with a uniform grid, it can be barycentrically interpolated from the nearest three pixels in the non-uniformly represented (modulated) features. This can be formulated as:
}
\begin{equation}
\begin{aligned}
&x' = w_{1} * x_{1} + w_{2} * x_{2} + w_{3} * x_{3}, 
\\
&\qquad \text{ s.t. } w_{1}+w_{2}+w_{3}=1.
\end{aligned}
\end{equation}
Where $x'$ is the interpolated pixel at the uniform grid $(i', j')$.
$x_{1}$, $x_{2}$, and $x_{3}$ are the nearest three pixels at $(i_1, j_1)$, $(i_2, j_2)$, and $(i_3, j_3)$ in the non-uniform modulated features.
$w_{1}$, $w_{2}$, and $w_{3}$ are the corresponding interpolation weights that can be solved by the equation below:
\begin{equation}
\begin{aligned}
&w_{1} = \frac{(j_2 - j_3)(i' - i_3) + (i_3-i_2)(j'-j_3)}
{(j_2 - j_3)(i_1 - i_3) + (i_3-i_2)(j_1-j_3)}, 
\\
&w_{2} = \frac{(j_3 - j_1)(i' - i_3) + (i_1-i_3)(j'-j_3)}
{(j_2 - j_3)(i_1 - i_3) + (i_3-i_2)(j_1-j_3)}, 
\\
& w_{3} = 1 - w_{1} - w_{2}.
\end{aligned}
\end{equation}

\vspace{+0.518mm}
{
\noindent{\bf Multi-scale relation.}
Non-uniform upsampling can effectively demodulate the feature but leaves two following issues unsolved:
1) {\it Lack of information interaction.}
Densely sampled areas typically contain richer information than sparsely sampled areas, but there is no explicit mechanism for information interaction between them to refine the prediction.
2) {\it Potential degradation caused by the sparsely sampled area.}
Due to non-uniform upsampling, in sparsely sampled areas, only a few wrong original pixels may be upsampled to a large area of wrong predictions, resulting in accuracy degradation.

To address these concerns, we propose cascaded Local Pixel Relation Modules (LPRM) to refine predictions during upsampling by exploiting multi-scale relations.
Each LPRM employs a $3\times 3$ convolutional layer with varying dilation to model pixel relations within a limited area. For flexibility, LPRM dynamically generates local pixel relation conditions based on the input compressed feature and then refines predictions using these conditions. This approach ensures adaptability and spatial-variability, allowing for desired flexibility.
To capture local information at multiple scales, the dilation in the latter LPRM is double that of the former one.

Specifically, as illustrated in Figure~\ref{fig:MSAU}, the first LPRM in MSAU takes compressed and non-uniform upsampled feature $\mathbf{X}^{comp}$ as input, and outputs local pixel relation $\mathbf{R}$ to refine the prediction:
\begin{equation}
\begin{aligned}
	\mathbf{R} = \text{Softmax}(Conv_{k\times k}^{d}(\mathbf{X}^{comp})),
\end{aligned}
\end{equation}
\begin{equation}
\begin{aligned}
\label{eq:refine}
	\mathbf{Y}'_{i, j} &= \sum_{p,q\in \Omega}
	\mathbf{R}_{i, j}^{p,q} \cdot \mathbf{Y}_{i+p, j+q},
\end{aligned}
\end{equation}
where $Conv_{k\times k}^{d}$ is convolution with dilation of $d$, and kernel size of $k$, 
$k=3$ is enough for both accuracy and efficiency.
{\color{diff}
The $\Omega$ indicates the neighbor area that $Conv_{k\times k}^{d}$ conditions on.
And $\mathbf{Y}$, $\mathbf{Y'}$ are prediction and refined prediction. 
The next LPRM further extracts the local pixel relation by using a convolution with larger dilation $d'$
}
\begin{equation}
\begin{aligned}
\label{eq:LPRM}
	\mathbf{R'} = \text{Softmax}(Conv_{k\times k}^{d'}(\mathbf{R})).
\end{aligned}
\end{equation}
The subsequent refinement follows the same process as Equation~\eqref{eq:refine}. In short, cascaded LPRMs iteratively refine the prediction according to Equations~\eqref{eq:LPRM} and \eqref{eq:refine}. As dilation increases, the LPRM in MSAU obtains wider receptive fields, enabling it to correct the prediction with the assistance of local information at multiple scales.
}

\subsection{Feature Frequency Analysis}

In this section, we analyze the frequency distribution of the feature map to assess the effectiveness of the proposed method using the Cityscapes~\cite{cityscapes2016} validation set.
Inspired by~\cite{2021deepfrequencyprinciple}, we utilize the Ratio Density Function (RDF) to characterize the power of each frequency.
RDF is defined as:
\begin{equation}
\begin{aligned}
\text{RDF}(\xi) = \frac{\partial \text{LFR}(\xi)}{\partial \xi},
\end{aligned}
\end{equation}
LFR, which stands for Low-Frequency Ratio, is defined as:
\begin{equation}
\begin{aligned}
\text{LFR}(\xi) = \frac{\sum \mathbb{I}_{\max(\left|\frac{k}{M}\right|, \left|\frac{l}{N}\right|)<\xi}\left|F(k,l)\right|^{2}}{\sum \left|F(k,l)\right|^{2}},
\end{aligned}
\end{equation}
{\color{diff}
where $\mathbb{I}_{\max(\left|\frac{k}{M}\right|, \left|\frac{l}{N}\right|)<\xi}$ is the indicator function that outputs 1 if $\max(\left|\frac{k}{M}\right|, \left|\frac{l}{N}\right|)<\xi$, and 0 otherwise.
Next, we employ this tool to analyze both the modulated and demodulated features.
}

{
\vspace{+0.518mm}
\noindent{\bf Modulated feature frequency analysis.}
We quantitatively analyze frequency in the feature map using RDF and show the results in Figure~\ref{fig:frequency_analysis}{\color{black}(a)}.
After modulation, compared with the original feature map, we observe that the power of frequencies larger than Nyquist frequency ($\xi > \frac{1}{4}$ for 2$\times$ downsampling) are largely reduced.
This indicates that ARS effectively modulates high frequencies to lower frequencies, thereby reducing the aliasing ratio and optimizing the feature frequency distribution.

Furthermore, we visualize the feature map and show typical results in Figure~\ref{fig:feature_freq_vis}.
When comparing the modulated feature in Figure~\ref{fig:feature_freq_vis}{\color{black}(b)} with the original feature map in Figure~\ref{fig:feature_freq_vis}{\color{black}(a)}, it becomes evident that the modulated feature exhibits a darker corner in the frequency spectra, indicating a reduction in high frequencies. This result is in line with our quantitative analysis.
}

{\color{diff}
\vspace{+0.518mm}
\noindent{\bf Demodulated feature frequency analysis.}
As shown in Figure~\ref{fig:frequency_analysis}{\color{black}(b)}, we further analyze the frequency components in the demodulated feature. Due to downsampling, high frequencies above the Nyquist frequency are either lost or aliased. Consequently, even after upsampling to the original size, the curve of the uniformly downsampled feature (light blue) in the high-frequency region experiences a significant reduction compared to the original feature (steel blue curve).
In contrast, the curve of our demodulated feature (light orange) is very closer to the original feature than the baseline upsampled feature, particularly for frequencies above the Nyquist frequency. This observation indicates that the proposed method effectively mitigates the aliasing effect, resulting in less information loss during downsampling and more efficient restoration of high frequencies during demodulation.
}

Moreover, we visualize the feature map and present representative results in Figure~\ref{fig:feature_freq_vis}. When comparing the demodulated feature in Figure~\ref{fig:feature_freq_vis}{\color{black}(d)} with the upsampled feature map in Figure~\ref{fig:feature_freq_vis}{\color{black}(c)}, it becomes apparent that the demodulated feature exhibits a brighter corner in the frequency spectrum, indicating the recovery of more high frequencies. This observation is consistent with our quantitative analysis. 
These results substantiate the superiority of the proposed method.

\section{Experimental Results}
We perform extensive experiments on Cityscapes~\cite{cityscapes2016}, Pascal Context~\cite{2014pascalcontext}, and ADE20K~\cite{ade20k}. We begin with an ablation study, followed by additional results for discussion, and finally conclude with a comparison to the state-of-the-art.

\subsection{Datasets and Metrics}
\vspace{+0.518mm}
\noindent{\bf Datasets.}
For our experiments, we utilize three challenging datasets: Cityscapes~\cite{cityscapes2016}, Pascal Context~\cite{2014pascalcontext}, and ADE20K~\cite{ade20k}.
Cityscapes~\cite{cityscapes2016} comprises 5,000 finely annotated images, each with dimensions of $2048 \times 1024$ pixels. The dataset is split into 2,975 images for the training set, 500 for the validation set, and 1,525 for the testing set. 
It contains 19 semantic categories for semantic segmentation tasks.
The Pascal Context~\cite{2014pascalcontext} contains 4,998 images for training and 5,105 images for validation/testing. 
Following~\cite{yu2020context, 2022pcaa}, without considering the background, we evaluate the performance on the most frequent 59 classes. 
The ADE20K~\cite{ade20k} is a challenging dataset that contains 150 semantic classes. 
It consists of 20,210, 2,000, and 3,352 images for the training, validation, and test sets. 





\vspace{+0.518mm}
\noindent\textbf{Metrics.} 
For semantic segmentation, consistent with~\cite{2016fcn, pspnet, 2020ccnet, 2021ocnet, 2022pcaa}, we evaluate the accuracy with mIoU (mean Intersection over Union). 
The GFLOP, parameter results are reported for computational cost. 

\subsection{Implementation Details}
By default, we use Stochastic Gradient Descent (SGD)~\cite{robbins1951stochastic} optimizer for training with momentum 0.9 and weight decay 0.0005.
Networks are trained for 40K iterations on Cityscapes~\cite{cityscapes2016} and Pascal Context~\cite{2014pascalcontext}. 
The initial learning rate is set to 0.01 for the Cityscapes, and 0.004 for the Pascal Context. 
Following~\cite{deeplabv3plus}, we employ the “poly” learning rate policy. 
The learning rate is multiplied by $(1-\frac{iter}{max\_iter})^{0.9}$.
To avoid over-fitting, we follow~\cite{pspnet, 2022pcaa} to adopt data augmentation strategies, including random cropping, random horizontal flipping, random photometric distortion, and random scaling. 
Crop size is $512 \times 1024$ for Cityscapes, $512 \times 512$ for the Pascal Context and ADE20K, $1024 \times 1024$ for COCO.
The batch size is 8 on Cityscapes, 16 on the Pascal Context, ADE20K and COCO. 
We set the weights of frequency modulation ($L_{\text{FM}}$) and semantic high-frequency loss ($L_{\text{SHF}}$) to 0.01 and 100, respectively.
For Mask2Former~\cite{2022mask2former}, and InternImage~\cite{2023internimage}, we use the same setting in their original papers.

\begin{figure}[t!]
\centering
\begin{tabular}{ccc}
\includegraphics[width=0.828\linewidth]{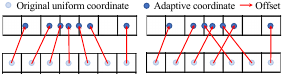}\\
\footnotesize
Adaptive Resampling \hspace{+8mm} Deformable Convolution
\end{tabular}
\vspace{-3.98mm}  
\caption{
Comparison of adaptive resampling and deformable convolution regarding coordinate sampling, illustrated in 1-Dimension.
}
\label{fig:deform}
\vspace{-3.98mm}
\end{figure}

{
\subsection{Comparison with State-of-the-arts}
In this section, we compare our approach with various state-of-the-art methods including low-pass pooling-based methods (Blur~\cite{2019makingshiftinvariant}, Adaptive blur~\cite{2020delving}, FLC~\cite{2022flc}), powerful deformable convolution~\cite{2019deformablev2}, 
and segmentation methods (CCNet~\cite{2020ccnet}, DeepLabV3+\cite{deeplabv3plus}, OCRNet\cite{2020ocrnet}, Segmentor~\cite{2021segmenter}, SETR~\cite{2021setr}, SegFormer~\cite{2021segformer}, HRFormer~\cite{hrformer}, MaskFormer~\cite{2021maskformer}, Mask2Former~\cite{2022mask2former}, and InternImage~\cite{2023internimage}).
These results demonstrate the advantages of the proposed methods. 
For example, our method outperforms the large InternImage-L/Mask2Former-Swin-L model~\cite{2023internimage} by a considerable margin of 0.8/0.6 mIoU on the challenging ADE20K~\cite{ade20k}.


\vspace{+0.518mm}
\noindent{\bf Comparison with low-pass pooling-based methods.}
Pooling-based methods alleviate aliasing by applying a low-pass filter to remove high frequencies but sacrifice fine details and textures, which are essential for accurate scene parsing and understanding.
Unlike these methods, our approach modulates high frequencies into lower frequencies and subsequently recovers them, instead of directly removing them.
When compared to pooling-based methods such as Blur~\cite{2019makingshiftinvariant}, Adaptive Blur~\cite{2020delving}, and FLC pooling~\cite{2022flc}, the proposed method outperforms them by a substantial margin of 0.7/0.6/0.9 mIoU (79.5 \emph{v.s.} 78.8, 78.9, 78.6), as shown in Table~\ref{tab:pooling_based}.
These results verify the effectiveness of the proposed strategy.

\vspace{+0.518mm}
\noindent{\bf Comparison with deformable convolution.}
The deformable convolution (DeformConv)~\cite{2017deformable, 2019deformablev2} is a powerful tool to model geometric transformations of objects and achieve great success on segmentation tasks.
Original deformable convolution (DeformConv)~\cite{2017deformable, 2019deformablev2} first predicts sampling offsets and then applies convolution with offsets.
Notably, DeformConv predicts offsets independently for each pixel carries the risk of disturbing the relative positions of pixels within the grid as illustrated in Figure~\ref{fig:deform}. 
For instance, background pixels might inadvertently sample object pixels, and objects could sample background pixels, potentially resulting in misclassification.

In contrast, the proposed ARS follows a different approach where it initially predicts an attention map and then generates sampling coordinates based on this map, using Equation~\eqref{eq:coordinates_mapping}. 
Each sampling coordinate generated is an attention-weighted average of neighboring pixel coordinates, considers neighboring pixels, and offers improved interpretability. 
Moreover, the coordinate mapping function Equation~\eqref{eq:coordinates_mapping} forms a monotonically increasing function. This unique characteristic ensures that the adaptive sampling coordinates maintain continuity. 
In contrast, DeformConv lacks this property, leading to potential discontinuities that may disturb the features, which is harmful.
Consequently, DeformConv displays limited improvements when applied to downsampling, as evidenced by Table~\ref{tab:dcn} (DeformConv 73.6 mIoU, compared to our ARS-DeformConv 74.7 mIoU). To counter the drawback of DeformConv, we extend the use of ARS to predict the sampling coordinates for the center of the kernel in DeformConv. This enhanced variant of DeformConv is referred to as ARS-DeformConv. 
The results, as presented in Table~\ref{tab:dcn}, indicate that DeformConv reaches 79.5 mIoU, and ARS-DeformConv provides an additional improvement of 0.8 mIoU over DeformConv, resulting in a cumulative enhancement of 80.3 mIoU.

\vspace{+0.518mm}
\noindent{\bf Comparison with state-of-the-art on ADE20K.}
Proofread: As shown in Table~\ref{tab:ade20k_main}, we compare the proposed method with state-of-the-art segmentation methods, including CCNet~\cite{2020ccnet}, DeepLabV3+~\cite{deeplabv3plus}, OCRNet~\cite{2020ocrnet}, CTNet~\cite{2021ctnet}, Segmentor~\cite{2021segmenter}, SETR~\cite{2021setr}, SegFormer~\cite{2021segformer}, HRFormer~\cite{hrformer}, MaskFormer~\cite{2021maskformer}, Mask2Former~\cite{2022mask2former}, Mask DINO~\cite{2023maskdino}, and Vision Adapter~\cite{2023vitadapter} on the challenging ADE20K~\cite{ade20k} dataset.
By incorporating SFM, we effectively outperform the large model UPerNet-InternImage-L by 0.8 mIoU (54.7 \emph{vs.} 53.9) and Mask2Former-Swin-L by 0.6 mIoU (56.7 \emph{vs.} 56.1), while introducing minor extra computational cost.
It is worth noting that with the same backbone of Swin-L, the proposed method outperforms the recent state-of-the-art Vision Adapter~\cite{2023vitadapter} (56.7 \emph{vs.} 53.4) with remarkably fewer parameters ($\sim$217M \emph{vs.} 364M).
Furthermore, the proposed method achieves higher results compared to the recent state-of-the-art Mask DINO~\cite{2023maskdino} (56.7 \emph{vs.} 56.6) with fewer parameters ($\sim$217M \emph{vs.} 223M).
These results verify the effectiveness of SFM.
}

\begin{table}[tb!]
\centering
\caption{
Comparison with low-pass pooling-based methods.
}
\vspace{-3.98mm}
\scalebox{0.98}{
\begin{tabular}{l|c|cc}
\toprule[1.28pt]
Method & mIoU & FLOPs & Params \\
\midrule
UPerNet-ResNet50~\cite{2018upernet} & 78.1 & 297.96G & 31.16M \\
\midrule
+Blur~\cite{2019makingshiftinvariant} & 78.8 & 298.49G & 31.16M \\
+AdaBlur~\cite{2020delving} & 78.9 & 336.56G & 32.32M \\
+FLC~\cite{2022flc} & 78.6 & 298.91G & 31.16M \\
\midrule
\rowcolor{mygray!58}
Ours &\bf 79.5 & 302.33G & 32.63M \\
\bottomrule[1.28pt]
\end{tabular}
}
\label{tab:pooling_based}
\vspace{-3.98mm}
\end{table}
\begin{table}[tb!]
\centering
\caption{
Comparison with DeformConv~\cite{2019deformablev2} on the Cityscapes validation set.
}
\vspace{-3.98mm}
\scalebox{0.988}{
\begin{tabular}{l|l|cccccccc}
\toprule[1.28pt]
Method & Downsample type & mIoU  \\
\midrule
\multirow{3}{*}{PCAA-R50-32$\times$~\cite{2022pcaa}}
&Strided-Conv (Baseline) & 73.3\\
&DeformConv~\cite{2019deformablev2} & 73.6 \\
&\cellcolor{mygray!58}ARS + Strided-Conv (Ours) & \cellcolor{mygray!58} \bf 74.7 (+1.1)\\
\midrule
\midrule
Method & Convolution type & mIoU  \\
\midrule
\multirow{3}{*}{PCAA-R50-8$\times$~\cite{2022pcaa}}  & Standard Conv (Baseline)& 78.5 \\
 & DeformConv~\cite{2019deformablev2} & 79.5 \\
 & \cellcolor{mygray!58}ARS-DeformConv (Ours) &\cellcolor{mygray!58}  \bf 80.3 (+0.8) \\
\bottomrule[1.28pt]
\end{tabular}
}
\label{tab:dcn}
\vspace{-3.98mm}
\end{table}
\begin{table}[tb!]
\color{diff}
\centering
\caption{
Comparison with state-of-the-art segmentation methods on ADE20K~\cite{ade20k}, results are from their original paper. 
Backbones pre-trained on ImageNet-22K are marked with $\dagger$. 
``*'' indicates using crop size of 640$\times$640 for training and testing.
The FLOPs results are calculated with an image size of 512$\times$512.
}
\vspace{-3.98mm}
\scalebox{0.6918}{
\begin{tabular}{l|l|cc|cccccc}
\toprule[1.28pt]
\multirow{2}{*}{Method} & \multirow{2}{*}{Backbone} & Params & FLOPs & \multicolumn{2}{c}{mIoU} \\
\cline{5-6}
& & (M) & (G) & SS & MS  \\
\midrule
DeepLabV3+{\color{gray}\tiny [ECCV2018]}~\cite{deeplabv3plus}&ResNet-101 &62.7  & 255.1 & 44.1 & - \\
CCNet{\color{gray}\tiny [TPAMI2020]}~\cite{2020ccnet} &ResNet-101 &68.9 & 278.4 & 45.2 & - \\
OCRNet{\color{gray}\tiny [ECCV2020]}~\cite{2020ocrnet} &HRNet-W48 &70.5  & 164.8 & 45.6 & - \\
CTNet{\color{gray}\tiny [TPAMI2022]}~\cite{2021ctnet} &ResNet-101 & 52.9 & 199.5 & 45.9 & - \\
\midrule
SFNet{\color{gray}\tiny [ECCV2020]}~\cite{2020semanticflow} &ResNet-101 &55.0 & 119.4 & - & 44.7 \\
AlignSeg{\color{gray}\tiny [TPAMI2021]}~\cite{2021alignseg} &ResNet-101 & -   & - & -  & 46.0 \\
IFA + ASPP{\color{gray}\tiny [ECCV2022]}~\cite{2022IFA} &ResNet-101 &64.3  & 72.0 & -  & 46.0 \\
\midrule
SETR-MLA{\color{gray}\tiny [CVPR2021]}~\cite{2021setr} & ViT-L$\dagger$ &310.6 & - & - & 50.3 \\
HRFormer{\color{gray}\tiny [NeurIPS2021]}~\cite{2021setr} & HRFormer-B &56.2 & 280.0 & 48.7 & 50.0 \\
\midrule
\multirow{2}{*}{SegFormer{\color{gray}\tiny [NeurIPS2021]}~\cite{2021segformer}}
& MiT-B4 & 64.1 & 95.7 & 50.3  & 51.1 \\
& MiT-B5* & 84.7 & 111.3 & 51.0 & 51.8 \\
\midrule
\multirow{2}{*}{MaskFormer{\color{gray}\tiny [NeurIPS2021]}~\cite{2021maskformer}}
& Swin-T & 42.0 & 55.0 & 46.7 & 48.8 \\
& Swin-L$\dagger$* & 212.0 & 375.0 &54.1 & 55.6 \\
\midrule
\multirow{2}{*}{InternImage{\color{gray}\tiny [CVPR2023]}~\cite{2023internimage}} 
& InternImage-T & 59.0 & 236.1 & 47.9 & 48.1 \\
& InternImage-L$\dagger$*& 255.6 & 404.2 & 53.9 & 54.1 \\
\midrule
\multirow{2}{*}{Mask2Former{\color{gray}\tiny [CVPR2022]}~\cite{2022mask2former}} 
& Swin-T & 47.0 & 74.0 & 47.7 & 49.6\\
& Swin-L$\dagger$* & 215.0 & 403.0 &56.1 & 57.3 \\
\midrule

Mask DINO{\color{gray}\tiny [CVPR2023]}~\cite{2023maskdino} &Swin-L$\dagger$* &223.0  & -  & 56.6 & -\\
Vision Adapter{\color{gray}\tiny [ICLR2023]}~\cite{2023vitadapter} &ViT-L$\dagger$* &363.8  & -  & 53.4 & 54.4\\

InceptionNeXt-B{\color{gray}\tiny [CVPR2024]}~\cite{2024inceptionnext} & InceptionNeXt-B & 115M & 1159G & - & 50.6 \\
PeLK-B{\color{gray}\tiny [CVPR2024]}~\cite{2024pelk} & PeLK-B & 126M & 1237G & 50.4 & - \\
MogaNet-L{\color{gray}\tiny [ICLR2024]}~\cite{2024moganet} & MogaNet-L & 113M & 1176G & 50.9 & - \\
ConvFormer-M36{\color{gray}\tiny [TPAMI2024]}~\cite{2024metaformer} &ConvFormer-M36 & 85M & 1113G & 51.3 & - \\
OverLoCK-B{\color{gray}\tiny [CVPR2025]}~\cite{2025overlock}& OverLoCK-B & 124M & 1202G & 51.7 & 52.3 \\
S.Mamba-B{\color{gray}\tiny [ICLR2025]}~\cite{2025spatialmamba}& S.Mamba-B & 127M & 1176G & 51.8 & 52.6 \\

\midrule
\rowcolor{mygray!58} 
\multirow{1}{*}{InternImage{\color{gray}\tiny [CVPR2023]}~\cite{2023internimage}} 
& InternImage-T & 59.6 & 239.4 & 49.2{\scriptsize (+1.3)} & 49.5 {\scriptsize (+1.5)} \\
\rowcolor{mygray!58} 
\multirow{1}{*}{+ SFM (Ours)}
& InternImage-L$\dagger$* & 256.6 & 408.0 &54.7 {\scriptsize (+0.8)}  &54.9 {\scriptsize (+0.8)} \\
\midrule
\rowcolor{mygray!58} 
\multirow{1}{*}{Mask2Former{\color{gray}\tiny [CVPR2022]}~\cite{2022mask2former}} 
& Swin-T & 47.4 & 75.5 & 49.1{\scriptsize (+1.4)} & 50.6{\scriptsize (+1.0)} \\
\rowcolor{mygray!58} 
+ SFM (Ours)
& Swin-L$\dagger$* & 216.8 & 407.8 &\bf 56.7{\scriptsize (+0.6)} &\bf 57.8{\scriptsize (+0.5)}\\
\bottomrule[1.28pt]
\end{tabular}
}
\label{tab:ade20k_main}
\vspace{-3.98mm}
\end{table}

\begin{table}[tb!]
\centering
\caption{
Top-1 accuracy results on CIFAR-100~\cite{2009cifar} test set and Tiny-ImageNet~\cite{2017tinyimagenet} validation set.
We simply replace the uniform downsampling in the model with adaptive resampling (ARS).
Computational cost is reported on $64\times 64$ size.
}
\vspace{-3.98mm}
\scalebox{0.88}{
\begin{tabular}{l|cc|c|c}
\toprule[1.28pt]
\multirow{1}{*}{Backbone} & \multirow{1}{*}{FLOPs} & \multirow{1}{*}{Parameters}  & CIFAR-100 & Tiny-ImageNet \\
\midrule
ResNet-50~\cite{resnet2016} & 5.26G & 23.71M & 81.36 & 74.31 \\
\rowcolor{mygray!58} 
+ ARS (Ours)& 5.29G & 25.10M &\bf 82.18 (+0.82) & \bf 74.98 (+0.67) \\
\midrule
ResNeXt-50~\cite{2017resnext} &5.46G &23.18M & 81.73 & 75.39 \\
\rowcolor{mygray!58} 
+ ARS (Ours)&5.49G &24.57M &\bf  82.41 (+0.68) & \bf 75.79 (+0.40) \\
\bottomrule[1.28pt]
\end{tabular}
}
\label{tab:classification}
\vspace{-3.98mm}
\end{table}

\begin{table}[tb!]
\centering
\caption{
FGSM adversarial training on CIFAR-10 using Preact-ResNet-18~\cite{2016identity} architecture (Baseline). We assess clean and robust accuracy, with higher values indicating better performance, against PGD~\cite{2016PGD} and AutoAttack~\cite{2020autoattack} on the test set with an $L_{\infty}$ norm and $\epsilon = 8/255$.
}
\vspace{-3.98mm}
\scalebox{0.918}{
\begin{tabular}{llccccc}
\toprule[1.28pt]
\multirow{2}{*}{Method} & \multirow{2}{*}{From}  & \multirow{2}{*}{Clean}  & PGD $L_{\infty}$ & AA $L_{\infty}$  \\
&  &  & $\epsilon = \frac{8}{255}$ & $\epsilon = \frac{8}{255}$  \\
\midrule
\multicolumn{5}{c}{CIFAR-10} \\
\midrule
Baseline & \it ECCV 2016   & 80.34   & 0.02 & 9.61 \\
\midrule
\rowcolor{mygray!58}
 + ARS (Ours) &  &\bf 80.64   &\bf 32.74 &\bf 27.95\\
\midrule
\midrule
Blur~\cite{2019makingshiftinvariant} &  \it ICML 2019 & 81.15 &35.09 & 33.52 \\
AdaBlur~\cite{2023depthadablur} &  \it NC 2023 &81.72 &35.17 & 28.74 \\
Wavelet~\cite{2021wavecnet} &  \it TIP 2021 &81.24 &33.57 &28.80\\
FLC~\cite{2022flc} & \it ECCV 2022 & 82.65 &35.17 & 29.79\\
\midrule
\rowcolor{mygray!58} 
FLC + ARS & Ours &\bf 83.10 & \bf 36.75   &\bf 30.59 \\ 


\midrule
\multicolumn{5}{c}{CIFAR-100} \\
\midrule
Baseline & \it ECCV 2016 & 46.60 &12.46 & 6.31\\
FLC~\cite{2022flc} & \it ECCV 2022 & 48.13 &13.57 & 6.70\\
\midrule
\rowcolor{mygray!58} 
FLC + ARS & Ours &\bf 48.39 & \bf 14.15   &\bf 7.07 \\  
\bottomrule[1.28pt]
\end{tabular}
}
\label{tab:robustness}
\vspace{-3.98mm}
\end{table}

\begin{table}[tb!]
\centering
\caption{
Extension to instance/panoptic segmentation.
All models are trained on COCO~\cite{mscoco2014} for 12 epochs.
\vspace{-3.98mm}
}
\scalebox{0.958}{
\begin{tabular}{l|l|ccccc}

\toprule[1.28pt]
\multirow{2}{*}{Task} &\multirow{2}{*}{Model} & AP$^{mask}$ & PQ \\
& &{COCO} \texttt{val} &{COCO} \texttt{val}\\
\midrule
\multirow{1}{*}{\it Instance}
&\multirow{1}{*}{Mask2Former-R50~\cite{2022mask2former}}
& 38.5   & -   \\
\multirow{1}{*}{\it Segmentation }
&\cellcolor{mygray!58} +SFM (Ours)
&\cellcolor{mygray!58}  \bf 39.2 (+0.7)   & \cellcolor{mygray!58}  -   \\
\midrule
\multirow{1}{*}{\it Panoptic } 
&\multirow{1}{*}{Mask2Former-R50~\cite{2022mask2former}}
& -  & 46.8    \\
\multirow{1}{*}{\it Segmentation}
& \cellcolor{mygray!58}  +SFM (Ours)
& \cellcolor{mygray!58} -    & \cellcolor{mygray!58} \bf 47.6 (+0.8)    \\
\bottomrule[1.28pt]
\end{tabular}
}
\label{tab:extension}
\vspace{-3.98mm}
\end{table}
\begin{figure}[t!]
\centering
\scalebox{0.99}{
\begin{tabular}{ccccc}
\hspace{-3mm}\includegraphics[width=0.99\linewidth]{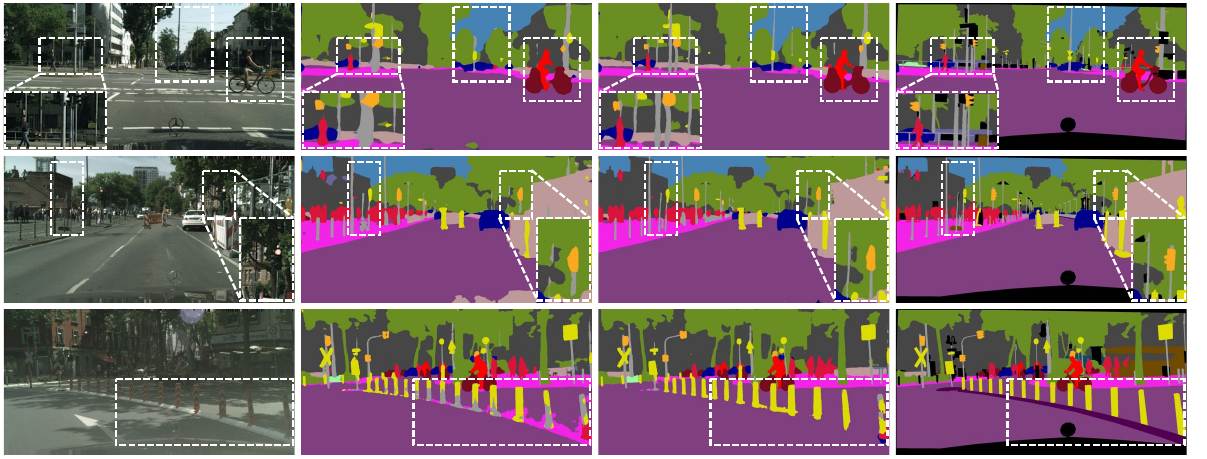} \\
\footnotesize
Image \hspace{+8mm}\quad w/o SFM \hspace{+8mm}\quad w/ SFM \hspace{+8mm}\quad Ground truth
\end{tabular}
}
\vspace{-3.98mm}    
\caption{
 Visualization on the Cityscapes validation set.
The white dashed boxes highlight the improved regions.
Our method enhances boundary precision and segmentation consistency.
}
\label{fig:vis_cityscapes}
\vspace{-3.98mm}
\end{figure}

{
\subsection{Extension to More Tasks}
In this section, we comprehensively evaluate the generalization ability of our method by extending it to various visual tasks, including image classification, robustness against adversarial attacks, and instance/panoptic segmentation.

\vspace{+0.518mm}
\noindent{\bf Image classification.}
Our evaluation encompasses two widely-used datasets: CIFAR-100~\cite{2009cifar} and Tiny-ImageNet~\cite{2017tinyimagenet}. 
To ensure a robust evaluation, we employ models~\cite{resnet2016,2017resnext} pre-trained on large ImageNet dataset~\cite{imagenet2015}. 
For fine-tuning, we iterate over 10 epochs with a batch size of 128. The learning rate follows a schedule of 0.05, with a decay ratio of 0.2 at 3, 6, and 9 epochs.
In contrast to semantic segmentation, image classification leans less heavily on spatial information and high-frequency components~\cite{2017dilated, 2020highfrequency}. 
This difference can be attributed to the widespread usage of global pooling~\cite{resnet2016,2017resnext} in classifiers, which only extracts the lowest frequency component of Discrete Cosine Transform (DCT)\cite{2021fcanet}. 
Consequently, the enhancement in classification accuracy, as demonstrated in Table~\ref{tab:classification}, is relatively modest compared to the gains observed in semantic segmentation.


\vspace{+0.518mm}
\noindent{\bf Robustness under adversarial attacks.}
Recent research~\cite{2020delving, 2022flc} has revealed the relationship between aliasing and model robustness.
This discovery has prompted the development of a series of pooling-based methods~\cite{2020delving, 2019makingshiftinvariant, 2021wavecnet, 2023depthadablur, 2022flc} that directly remove the high-frequency components responsible for aliasing.
Our approach introduces a novel and alternative solution: by modulating high-frequency signals to lower frequencies, we effectively mitigate aliasing, leading to enhanced model resistance against adversarial attacks.
The results presented in Table~\ref{tab:robustness} demonstrate its improvement on the Preact-ResNet-18 for both clean images and under adversarial attacks~\cite{2016PGD, 2020autoattack}. Moreover, it can combine and further improve state-of-the-art methods like FLC~\cite{2022flc}, outperforming previous methods such as Blur~\cite{2019makingshiftinvariant}, AdaBlur~\cite{2020delving}, Wavelet~\cite{2021wavecnet}, and FLC pooling~\cite{2022flc}. These results highlight the superiority of our ARS.

\vspace{+0.518mm}
\noindent{\bf Extension to instance/panoptic segmentation.}
We have expanded the application of our method to the domain of Instance/Panoptic Segmentation. 
We conduct experiments on the 2017 data splits of COCO dataset~\cite{mscoco2014}.
For instance segmentation, we use AP (average precision averaged over categories and IoU thresholds) as the primary metric.
For, panoptic segmentation, we use PQ (panoptic quality) as the default metric to measure panoptic segmentation performance~\cite{kirillov2019panoptic}.
As shown in Table~\ref{tab:extension}, our SFN enhances the performance of the state-of-the-art mask2former model with improvements of 0.7 and 0.8 for instance and panoptic segmentation, respectively. This result not only validates the effectiveness of our approach but also highlights its potential for generalization across various computer vision tasks.
}

\begin{figure}[t!]
\centering
\begin{tabular}{c}
\includegraphics[width=0.918\linewidth]{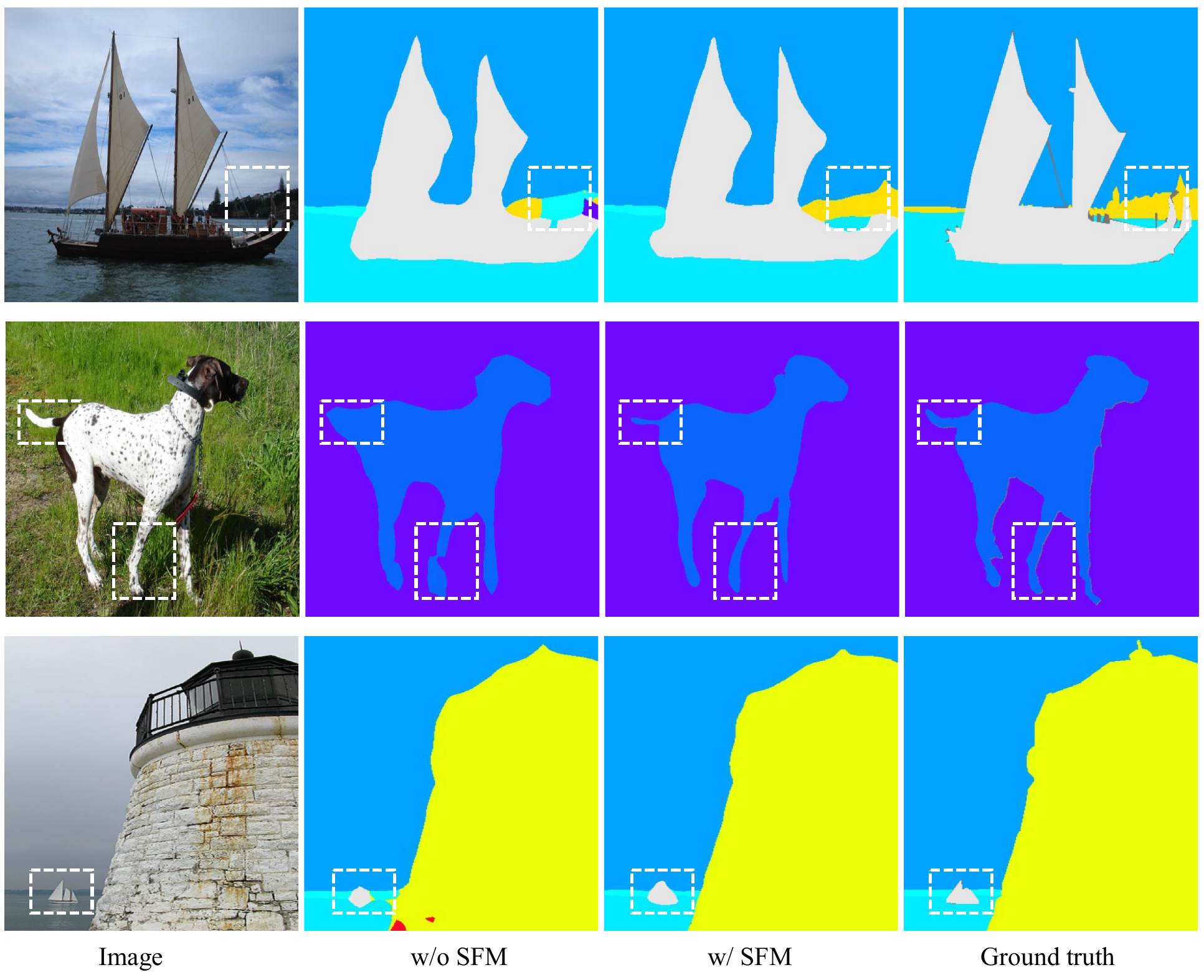} \\
\end{tabular}
\vspace{-3.98mm}    
\caption{
Visualized results on Pascal Context~\cite{2014pascalcontext} validation set.  
}
\label{fig:vis_pascal_context_d32}
\vspace{-3.98mm}
\end{figure}

\begin{figure}[t!]
\centering
\begin{tabular}{c}
\includegraphics[width=0.918\linewidth]{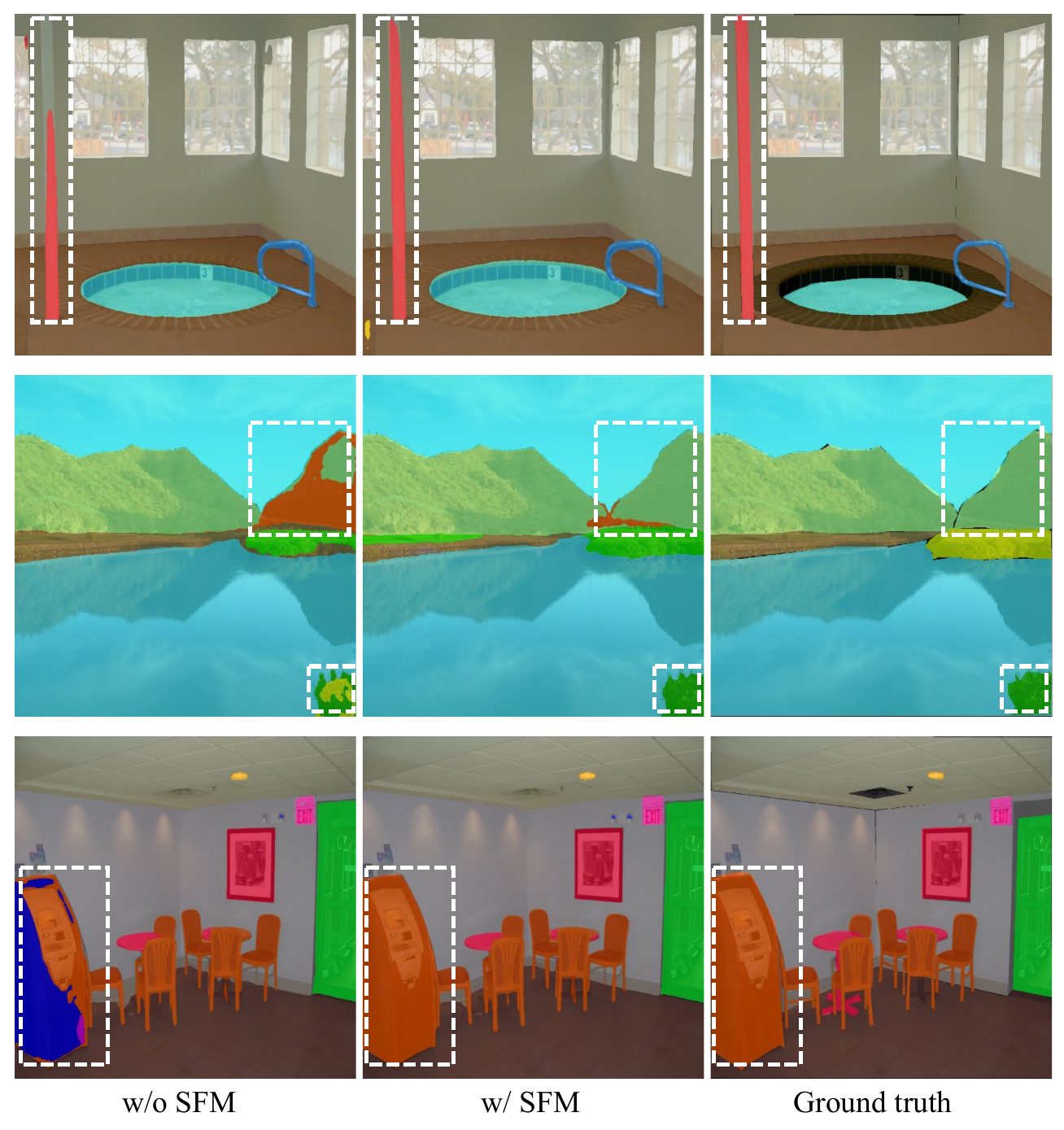} \\
\footnotesize
\end{tabular}
\vspace{-5mm}     
\caption{
Visualized results on ADE20K~\cite{ade20k} dataset.
}
\label{fig:vis_ade20k}
\vspace{-2.98mm}
\end{figure}

\begin{table*}[t!]
\centering
\caption{
Results of combining different state-of-the-art models with varying downsampling strides on the Cityscapes~\cite{cityscapes2016} validation set.
}
\vspace{-3.98mm}
\scalebox{0.8918}{
\begin{tabular}{c|c||ccc||ccc||ccc}
\toprule[1.28pt]
\multirow{2}{*}{Method} & \multirow{2}{*}{From} & \multicolumn{3}{c||}{Stride = 32 (ResNet-32$\times$~\cite{resnet2016})} & \multicolumn{3}{c||}{Stride  = 16 (Dilated ResNet-16$\times$~\cite{2017dilated})} & \multicolumn{3}{c}{Stride = 8 (Dilated ResNet-8$\times$~\cite{2017dilated})} \\
\cline{3-11}
&  & GFLOPs & Parameters & mIoU & GFLOPs & Parameters & mIoU & GFLOPs & Parameters & mIoU \\
\midrule
FCN~\cite{2016fcn} &\it ICCV & 229.65 & 49.50M & 71.2 & 463.31 & 49.50M & 73.0 & 1581.80 & 49.50M & 74.7 \\
\rowcolor{mygray!58}
+SFM (Ours) &\it 2015 & 231.06 & 50.93M & \bf 76.2 (+5.0) & 464.90 & 49.88M &\bf 77.8 (+4.8) & 1584.27 & 49.61M &\bf 78.7 (+4.0) \\
\midrule
PSPNet~\cite{pspnet}&\it ICCV & 220.06 & 48.98M & 72.7 & 424.77 & 48.98M &76.8 & 1427.47 & 48.98M & 78.4 \\
\rowcolor{mygray!58}
+SMF (Ours) &\it 2017 & 221.46 &50.41M & \bf 76.0 (+3.3) & 426.36 &49.35M &\bf 78.1 (+1.3) & 1429.94 &49.09M &\bf 79.5 (+1.1) \\ 
\midrule
CCNet~\cite{2020ccnet} &\it TPAMI & 231.00 & 49.83M & 73.3 & 468.69 & 49.83M & 75.6 & 1603.32 & 49.83M&77.8 \\
\rowcolor{mygray!58}
+SFM (Ours)&\it 2020 & 232.41 &51.26M & \bf 76.3 (+3.0)& 470.28 &50.20M&\bf 78.5 (+2.9) & 1605.79 &49.94M&\bf 79.3 (+1.5) \\
\midrule
OCNet~\cite{2021ocnet}& \it IJCV & 205.50 &37.71M & 72.5 & 366.71 &37.71M & 76.6 & 1195.37 &37.71M &78.5 \\
\rowcolor{mygray!58}
+SFM (Ours) &\it 2021 & 206.91 & 39.14M & \bf 75.8 (+3.3)& 368.26 & 38.08M &\bf 78.0 (+1.4) & 1197.85 & 37.82M &\bf 79.7 (+1.2) \\
\midrule
PCAA~\cite{2022pcaa} &\it CVPR & 230.37 & 50.30M & 73.3 & 465.71 & 50.30M &76.9 & 1590.89 & 50.30M &78.5 \\
\rowcolor{mygray!58}
+SFM (Ours)&\it 2022 & 231.78 &51.73M &\bf 76.8 (+3.5) & 467.29 &50.67M &\bf 77.9 (+1.0) & 1593.36 &50.41M &\bf 79.6 (+1.0) \\
\bottomrule[1.28pt]
\end{tabular}
}
\label{tab:main_cityscapes}
\vspace{-3.98mm}
\end{table*}

\begin{table}[t!]
\centering
\caption{
Results on the Pascal Context~\cite{2014pascalcontext} test set.
The backbone is ResNet-50 with a downsampling stride of $32\times$.
}
\vspace{-3.98mm}
\scalebox{0.98}{
\begin{tabular}{l|cccccc}
\toprule[1.28pt]
Method &FCN~\cite{2016fcn} & PSPNet~\cite{pspnet} & CCNet~\cite{2020ccnet} & OCNet~\cite{2021ocnet}\\
\midrule
Vanilla &43.8 &48.0 &47.7 & 47.9  \\
\rowcolor{mygray!58}
Ours & \bf 46.0 (+2.2) &\bf 49.7 (+1.7) &\bf 49.1 (+1.4) &\bf 49.9 (+2.0) \\
\bottomrule[1.28pt]
\end{tabular}
}
\label{tab:main_pascalcontext}
\vspace{-3.98mm}
\end{table}

\begin{table*}[tb!]
\caption{
Boundary accuracy and error analysis.
The $\uparrow$/$\downarrow$ indicates that higher/lower values are better. 
Results are reported on Cityscapes val~\cite{cityscapes2016}.
}
\vspace{-3.98mm}
\label{tab:boundary_analysis}
\centering
\scalebox{0.8918}{
\begin{tabular}{l|ll|lll|llll}
\toprule[1.28pt]
\multirow{2}{*}{Method} & \multicolumn{2}{c|}{Overall metric} & \multicolumn{3}{c|}{Boundary metric~\cite{gatedscnn2019, 2021boundaryiou}} & \multicolumn{3}{c}{Aliasing error metric~\cite{chen2023semantic}}\\
& {mIoU$\uparrow$} & {F-score$\uparrow$} & {BF-score$\uparrow$} & {BIoU$\uparrow$} & {BAcc$\uparrow$} & {FErr$\downarrow$} & {MErr$\downarrow$} & {DErr$\downarrow$} \\
\midrule
ResNet-32$\times$~\cite{resnet2016} & 72.7 & 83.1 & 68.4 & 53.2 & 65.8 & 36.8 & 35.8 & 39.7 \\
\rowcolor{mygray!58}
+SFM (Ours) 
&\bf 76.0{\scriptsize (+3.3)} 
&\bf 85.6{\scriptsize (+2.5)}
&\bf 71.9{\scriptsize (+3.5)}
&\bf 57.3{\scriptsize (+4.1)}
&\bf 70.0{\scriptsize (+4.2)}
&\bf 31.9{\scriptsize (-4.9)}
&\bf 30.8{\scriptsize (-5.0)}
&\bf 33.3{\scriptsize (-6.4)}
\\
\midrule
Dilated ResNet-16$\times$~\cite{2017dilated} & 76.8 & 86.2 & 73.2 & 58.9 & 71.5 & 30.6 & 29.0 & 31.7 \\
\rowcolor{mygray!58}
+SFM (Ours) 
&\bf 78.1{\scriptsize (+1.3)} 
&\bf 87.0{\scriptsize (+0.8)} 
&\bf 75.2{\scriptsize (+2.0)} 
&\bf 61.3{\scriptsize (+2.4)} 
&\bf 73.2{\scriptsize (+1.7)} 
&\bf 28.0{\scriptsize (-2.6)} 
&\bf 27.1{\scriptsize (-1.9)} 
&\bf 28.5{\scriptsize (-3.2)} \\
\midrule
Dilated ResNet-8$\times$~\cite{2017dilated} 
& 78.4 & 86.9 & 75.2 & 61.4 & 72.8 & 28.5 & 27.5 & 29.1 \\
\rowcolor{mygray!58}
+SFM (Ours) 
&\bf 79.5{\scriptsize (+1.1)}
&\bf 87.8{\scriptsize (+0.9)}
&\bf 76.3{\scriptsize (+1.1)}
&\bf 62.7{\scriptsize (+1.3)}
&\bf 74.2{\scriptsize (+1.4)}
&\bf 27.0{\scriptsize (-0.5)}
&\bf 25.9{\scriptsize (-2.4)}
&\bf 27.3{\scriptsize (-2.2)}\\
\bottomrule[1.28pt]
\end{tabular}}
\vspace{-3.98mm}
\end{table*}

\begin{table*}[tb!]
\centering
\caption{
Results of combinations with different backbone (ResNet~\cite{resnet2016}, Swin~\cite{2021swin}, CNN-based ConvNeXt~\cite{2022convnet}, and InternImage~\cite{2023internimage}) and segmentator architectures (FCN-based PSPNet~\cite{pspnet}, and FPN-based~\cite{2017feature} Mask2Former and UPerNet~\cite{2018upernet}).
}
\vspace{-3.98mm}
\scalebox{0.98}{
\begin{tabular}{l|c|cc|cc|cccc}
\toprule[1.28pt]
\multirow{2}{*}{FCN-based~\cite{2016fcn} architectures}  
& \multirow{2}{*}{From}
& \multicolumn{2}{c|}{GFLOPs (1024$\times$2048)} & \multicolumn{2}{c|}{Parameters} & \multicolumn{2}{c}{mIoU ({Cityscapes} \texttt{val})} \\
\cline{3-8}
& & Vanilla & Ours & Vanilla & Ours & Vanilla & Ours   \\
\midrule
ResNet-50-C~\cite{2019bagoftrick} &\multirow{2}{*}{\it CVPR 2016}  & 220.06 & 221.46 &48.98M &50.41M & 72.7 &\bf 76.0 (+3.3)\\
ResNet-101-C~\cite{2019bagoftrick} & &375.42 &376.83 & 67.97M & 69.40M & 73.8 &\bf 76.6 (+2.8) \\
\midrule
Swin-T~\cite{2021swin} &\multirow{2}{*}{\it CVPR 2021} & 194.20 & 195.28 &31.10M &31.32M &72.2 &\bf 74.6 (+2.4) \\
Swin-S~\cite{2021swin} & &376.21 &377.29 & 52.42M & 52.64M & 74.9 &\bf 76.8 +(1.9)\\
\midrule
ConvNeXt-T~\cite{2022convnet} &\multirow{2}{*}{\it CVPR 2022} & 191.76 &192.84 &31.40M &31.62M &72.8 &\bf 75.0 (+2.2)\\
ConvNeXt-S~\cite{2022convnet} & &368.54 &369.62 &53.04M &53.26M &74.0 &\bf 76.0 (+2.0) \\
\midrule
\midrule
\multirow{2}{*}{FPN-based\cite{2017feature} architectures} 
& \multirow{2}{*}{From}
& \multicolumn{2}{c|}{GFLOPs (512$\times$512)} & \multicolumn{2}{c|}{Parameters} & \multicolumn{2}{c}{mIoU ({ADE20K} \texttt{val})} \\
\cline{3-8}
& & Vanilla & Ours & Vanilla & Ours & Vanilla & Ours   \\
\midrule
Mask2Former-Swin-T~\cite{2022mask2former} &\it CVPR 2022 &74.00 &75.50 & 47.00M & 47.40M & 47.7 & \bf 49.2 (+1.5) \\
UPerNet-InternImage-T~\cite{2023internimage} &\it CVPR 2023  &236.13 & 239.35 &59.00M & 59.61M & 47.9 &\bf 49.3 (+1.4) \\
\bottomrule[1.28pt]
\end{tabular}
}
\label{tab:main_backbone}
\vspace{-5.18mm}
\end{table*}

\begin{table}[t!]
\caption{
Combination with shrunk version of SegViT~\cite{2022segvit} (naive shrunk) on the ADE20K dataset. The GFLOPs are measured at single scale inference with a crop size of 512$\times$512 on ViT-Base backbone.
}
\vspace{-3.98mm}
\label{tab:segvit}
\centering
\scalebox{0.918}{
\begin{tabular}{l|l}
\toprule[1.28pt]
Method & mIoU  \\
\midrule
DINOv2-base (shrunk) {\color{gray}\tiny [TMLR2023]}~\cite{2023dinov2} & 43.3 \\
\rowcolor{mygray!58}
DINOv2-base~\cite{2022segvit} + SFM (Ours) & \bf 44.8 {\scriptsize (+1.5)}\\
\midrule
SegViT-base (shrunk) {\color{gray} \tiny [NeurIPS2022]}~\cite{2022segvit} & 46.9 \\
\rowcolor{mygray!58}
SegViT-base~\cite{2022segvit} + SFM (Ours) & \bf 48.1 {\scriptsize (+1.2)}\\
\bottomrule[1.28pt]
\end{tabular}}
\vspace{-3.98mm}
\end{table}

\begin{table}[tb!]
\centering
\caption{
Comparison with image adaptive downsampler for efficient low-resolution semantic segmentation on the Cityscapes validation set~\cite{cityscapes2016}.
}
\vspace{-3.98mm}
\scalebox{0.98}{
\begin{tabular}{llccccc}
\toprule[1.28pt]
Method  & From  & GFLOPs & mIoU \\ 
\midrule
ES~\cite{2019efficientseg}  & \it ICCV 2019   & 44.2   & 54.0 \\
LDS~\cite{2022learningtodown} & \it ICLR 2022     & 50.0   & 55.0 \\
LZU~\cite{2023lzu} & \it CVPR 2023 & 45.0   & 55.0 \\ 
\midrule
\rowcolor{mygray!58}
SFM     &  Ours & \bf 15.3   &\bf 55.2 \\ 
\bottomrule[1.28pt]
\end{tabular}
}
\label{tab:low_resolution}
\vspace{-3.98mm}
\end{table}

\begin{table}[tb!]
\caption{
Comparison with the CondConv~\cite{2019condconv} and learnable upsampling strategy DySample~\cite{2023dysample}. The results are reported on the Cityscapes validation set~\cite{cityscapes2016}.
AR indicates aliasing ratio.
}
\vspace{-3.98mm}
\label{tab:condconv_dysample}
\centering
\scalebox{0.78}{
\begin{tabular}{l|c|c|l|l}
\toprule[1.28pt]
\multirow{2}{*}{Method} & Params. & FLOPs & \multirow{2}{*}{mIoU$\uparrow$} & \multirow{2}{*}{AR$\downarrow$}  \\
 & (M) & (G) & \\
\midrule
ResNet-50 & 48.98 & 220.06 & 72.7 & 42.7\%\\
\midrule
+ CondConv{\color{gray}\tiny [NeuIPS2019]}~\cite{2019condconv} & 82.62  & 220.07 & 73.8 {\scriptsize (+1.1)} & 43.8\% \\
+ DySample{\color{gray}\tiny [ICCV2023]}~\cite{2023dysample} & 49.50  &222.39  & 73.5 {\scriptsize (+0.8)}  & 42.8\%\\
\rowcolor{mygray!28}
\midrule
+ SFM (Ours)& 50.41  & 221.46 &\underline{76.0 {\scriptsize (+3.3)}} & 34.2\% \\
\rowcolor{mygray!58}
+ CondConv~\cite{2019condconv} + SFM (Ours) & 84.06 & 221.48 &\bf 76.8 {\scriptsize (+4.1)} & 34.0\%\\
\bottomrule[1.28pt]
\end{tabular}}
\vspace{-3.98mm}
\end{table}
\begin{table}[tb!]
\caption{
Ablation study for the proposed Adaptive Resampling (ARS) and Multi-Scale Adaptive Upsampling (MSAU).
Results are reported on the Cityscapes validation set.
Pyramid spatial pooling~\cite{pspnet} is used as the segmentation head.
}
\vspace{-3.98mm}
\label{tab:correlation}
\centering
\scalebox{0.918}{
\begin{tabular}{l|ll}
\toprule[1.28pt]
Method &mIoU \\
\midrule
ResNet-50 & 72.7 \\
\midrule
+ ARS  & 72.2 {\scriptsize (-0.5)}\\
+ MSAU & 74.0 {\scriptsize (+1.3)}\\
+ ARS + DySample~\cite{2023dysample} & 72.5 {\scriptsize (-0.2)}\\
\rowcolor{mygray!58}
+ ARS + MSAU (SFM) & \bf 76.0 {\scriptsize (+3.3)}\\
\bottomrule[1.28pt]
\end{tabular}}
\vspace{-3.98mm}
\end{table}

\begin{table*}[h!]
\centering
\caption{
\small
Comparison and combination with upsampling/segmentation refinement methods on the Cityscapes dataset~\cite{cityscapes2016} validation set.
}
\vspace{-3.98mm}
\scalebox{0.8288}{
\hspace{-5.18mm}
\begin{tabular}{l|c|ccccc|cc}
\toprule[1.28pt]
\multirow{2}{*}{Method}  & DeepLabv3   & +DenseCRF & +GUM & +DUsampling & +PointRend & +SegFix & +SFM & +SegFix \\
& {\color{gray}\scriptsize [arXiv2017]}~\cite{deeplabv3} & {\color{gray}\scriptsize [NeurIPS2011]}~\cite{2011densecrf}  & {\color{gray}\scriptsize [BMVC2018]}~\cite{2018gun} & {\color{gray}\scriptsize [CVPR2019]}~\cite{dupsample} & {\color{gray}\scriptsize [CVPR2020]}~\cite{pointsupervised} & {\color{gray}\scriptsize [ECCV2020]}~\cite{2020segfix} & (Ours) & +SFM (Ours)\\
\midrule
mIoU & 79.5 & 79.7 & 79.8 & 79.9 & 80.2 & 80.5 & \bf 81.0 & \bf 81.2 \\
\bottomrule[1.28pt]
\end{tabular}
}
\label{tab:segrefine}
\vspace{-5.18mm}
\end{table*}

\subsection{Discusion}
\label{sec:ablation}
{\color{diff}
In this section, we present additional results to comprehensively evaluate the proposed method across various settings on several prominent semantic segmentation benchmarks, including Cityscapes~\cite{cityscapes2016}, Pascal Context~\cite{2014pascalcontext}, and ADE20k~\cite{ade20k}.
We conduct an extensive evaluation of the proposed method, considering the following:
\begin{itemize}
\item Different segmentation heads, including FCN~\cite{2016fcn}, PSPNet~\cite{pspnet}, CCNet~\cite{2020ccnet}, OCNet~\cite{2021ocnet}, and PCAA~\cite{2022pcaa}.
\item Various downsampling strides (8$\times$, 16$\times$, and 32$\times$).
\item Detailed quantitative boundary analysis and error caused by aliasing are provided. 
\item Diverse backbones/architectures, such as widely-used ResNet~\cite{resnet2016}, Transformer-based Swin-Transformer~\cite{2021swin}, CNN-based ConvNeXt~\cite{2022convnet}, and InternImage~\cite{2023internimage}. In addition to the FCN-based method~\cite{pspnet}, we also evaluate using FPN-based UPerNet~\cite{2018upernet} and Mask2Former~\cite{2022mask2former} as segmentation architectures. 
\item ViT-based architectures are also discussed.
\item Recent state-of-the-art methods including InceptionNeXt~\cite{2024inceptionnext}, PeLK~\cite{2024pelk}, MogaNet~\cite{2024moganet}, ConvFormer~\cite{2024metaformer}, S.Mamba-B~\cite{2025spatialmamba}, and OverLoCK~\cite{2025overlock}.
\item Additionally, we provide visualized results for verification.
\end{itemize}

These comprehensive quantitative and qualitative evaluations demonstrate that the proposed method consistently brings improvements across various segmentation heads, backbones, and architectures while incurring minimal increases in computation and parameters. 
}

\vspace{+0.518mm}
\noindent{\bf Combining with different segmentation heads.}
We conduct experiments with the latest semantic segmentation methods, \ie, FCN~\cite{2016fcn}, PSPNet~\cite{pspnet}, CCNet~\cite{2020ccnet}, OCNet~\cite{2021ocnet} and PCAA~\cite{2022pcaa}.
As shown in Tables~\ref{tab:main_cityscapes} and~\ref{tab:main_pascalcontext}, with downsampling stride of $32\times$, our method consistently improves vanilla models by 3.0$\sim$5.0 mIoU on Cityscapes~\cite{cityscapes2016} and 1.4$\sim$2.2 mIoU on Pascal Context~\cite{2014pascalcontext}.

\vspace{+0.518mm}
\noindent{\bf Different downsampling strides.}
We also use dilated convolution~\cite{2017dilated} to experiment with three common downsampling strides, as shown in Table~\ref{tab:main_cityscapes}.
Specifically, we apply ARS before the last 3, 2, and 1 downsampling operations in models for 32$\times$, 16$\times$, and 8$\times$ downsampling strides, respectively.
The 8$\times$ downsampling stride setting has fewer downsampling operations, which means a slighter aliasing problem.
Thus, the performance improvement in the 8$\times$ setting is relatively less than $32\times$ downsampling stride (3.0$\sim$5.0 mIoU \emph{v.s.} 1.0$\sim$4.0 mIoU).
Besides, our method gets similar or better results compared with higher sampling rate setting, \eg, our 32$\times$ FCN~\cite{2016fcn} (76.2 mIoU, 230 GFLOPs) achieves 1.5 higher mIoU than vanilla 8$\times$ FCN (74.7 mIoU, 1582 GFLOPs).

\vspace{+0.518mm}
\noindent{\bf Boundary \& aliasing error rate analysis.}
Here, we have incorporated additional metrics, including the boundary F-score (BF-score), boundary IoU (BIoU)\cite{2021boundaryiou}, and boundary accuracy (BAcc), into our evaluation framework. 
{\color{diff}
Moreover, following~\cite{chen2023semantic}, we utilize metrics designed to analyze specific types of boundary errors potentially exacerbated by aliasing: false responses (FErr), merging mistakes (MErr), and displacements (DErr). Together, these boundary-specific metrics (BF-score, BIoU, BAcc, FErr, MErr, DErr) allow for a fine-grained evaluation of how well models handle critical boundary regions and the extent to which aliasing might be contributing to errors.
}

Specifically, in Table~\ref{tab:boundary_analysis}, we showcase the effectiveness of the SFM method across different downsampling strides. 
SFM demonstrates notable enhancements in boundary-related metrics such as BF-score, BIoU, and BAcc. For the ResNet-32$\times$ model, there are improvements of +3.5 in BF-score, +4.1 in BIoU, and +4.2 in BAcc, which is remarkable. 
Similarly, for the Dilated ResNet-16/8$\times$, improvements of +2.0/1.1 in BF-score, +2.4/1.3 in BIoU, and +1.7/1.4 in BAcc are considerable.
Additionally, SFM results in substantial reductions in error rates caused by aliasing, including FErr, MErr, and DErr. For instance, on the ResNet-32$\times$ model, FErr, MErr, and DErr decreased by 4.9, 5.0, and 6.4 points, respectively. 
These results indicate that the proposed SFM effectively reduces errors caused by aliasing, improving the accuracy and consistency of boundary segmentation.

\vspace{+0.518mm}
\noindent{\bf Combining with diverse backbones/architectures.}
We further choose ResNet~\cite{resnet2016}, Swin Transformer~\cite{2021swin}, and ConvNeXt~\cite{2022convnet} to comprehensively verify the adaptability of the proposed method.
The ResNet is a classic and widely used backbone, while Swin Transformer and ConvNeXt are the latest typical transformer-based and CNN-based backbones, respectively.
As shown in Table~\ref{tab:main_backbone}, the proposed method steadily improves the accuracy of these backbones, which means the proposed method generalizes well.
Besides, the proposed method helps shallow ResNet-50 and ConvNeXt-T outperform deeper ResNet-101 and ConvNeXt-S by a margin of 1.0$\sim$2.2\%.
These results demonstrate the effectiveness of our proposed method.

Additionally, we also apply the proposed method with FPN-based architecture including UPerNet~\cite{2018upernet} and Mask2Former~\cite{2022mask2former}, utilizing Swin Transformer~\cite{2021swin} and InternImage~\cite{2023internimage} as backbones.
FPN-based models utilize the lateral pathway to fuse high-frequency information in low-level features, thereby mitigating the high-frequency loss caused by downsampling. 
Despite this, on the challenging ADE20K~\cite{ade20k} dataset, our method achieves improvements of +1.5/1.4 mIoU for Mask2Former-Swin-T/UPerNet-InternImage-T as shown in Table~\ref{tab:main_backbone}.

{
\vspace{+0.518mm}
\noindent{\bf Combining with ViT structure.} 
Our proposed methods address the aliasing artifacts stemming from downsampling, but standard ViT-based models~\cite{2020vit} do not involve downsampling operations.
To combine with the ViT structure, we have conducted experiments on DINOv2~\cite{2023dinov2} and SegViT~\cite{2022segvit}, which employ a ``shrunk" downsampling strategy as described in~\cite{2022segvit}.
As shown in Table~\ref{tab:segvit}, we have achieved promising improvements.
Specifically, when incorporating SFM with DINOv2~\cite{2023dinov2}, a remarkable +1.5 mIoU improvement is achieved in comparison to the ``shrunk" version of DINOv2~\cite{2023dinov2}. Additionally, our integration of SFM with SegViT-Base~\cite{2022segvit} yields a substantial +1.2 mIoU increase over the ``shrunk" variant of SegViT~\cite{2022segvit}.
}

\vspace{+0.518mm}
\noindent{\bf Visualization.}
We illustrate visualized results on Cityscapes~\cite{cityscapes2016}, Pascal Context~\cite{2014pascalcontext}, and ADE20K~\cite{ade20k} in Figures~\ref{fig:vis_cityscapes},~\ref{fig:vis_pascal_context_d32}, and~\ref{fig:vis_ade20k}, respectively.
These visualization results further confirm that the proposed method enhances boundary precision and segmentation consistency. 
Moreover, the feature visualization in Figure~\ref{fig:featurevis2} demonstrates that the SFM framework effectively preserves semantic responses to objects during downsampling and recovers upsampled feature maps with rich fine details. 
These visualization results are consistent with the quantitative results.

\begin{table}[tb!]
\centering
\caption{
Ablation study for the numbers of Adaptive ReSampling (ARS).
The three ARS blocks are at $\text{stage1}\rightarrow \text{stage2}$, $\text{stage2}\rightarrow \text{stage3}$, and $\text{stage3}\rightarrow \text{stage4}$ of the ResNet~\cite{resnet2016}.
We employ non-uniform upsampling to reverse/demodulate the feature modulation to obtain the final prediction by default.
}
\vspace{-3.98mm}
\scalebox{0.8518}{
\begin{tabular}{c|ccc|cccc}
\toprule[1.28pt]
\#ARS &$\text{stage1}\rightarrow \text{stage2}$ & $\text{stage2}\rightarrow \text{stage3}$ & $\text{stage3}\rightarrow \text{stage4}$ & mIoU \\
\midrule
0 & &  &  & 72.7 \\
\midrule
1& &  & \checkmark & 73.9 (+1.2) \\
2& & \checkmark & \checkmark & 74.2 (+1.5)\\
3& \checkmark & \checkmark & \checkmark &\bf 74.7 (+2.0)\\
\bottomrule[1.28pt]
\end{tabular}
}
\label{tab:ars_number}
\vspace{-3.98mm}
\end{table}

\begin{table}[tb!]
\centering
\caption{
Ablation studies for attention map generation. `+ $L$' indicates using the frequency modulation and semantic high-frequency loss to supervise the learning of the attention map generator.
DAConv is the difference-aware convolution.
LPRM is not applied, and we only use non-uniform upsampling for demodulation.
}
\vspace{-3.98mm}
\scalebox{0.828}{
\begin{tabular}{c|l|cc|c}
\toprule[1.28pt]
Sampling & Method & GFLOPs & Parameters &mIoU \\
\midrule
\multirow{1}{*}{w/o SFM}
&Vanilla & 220.06 & 48.98M & 72.7\\
\midrule
\multirow{7}{*}{w/ SFM}
&Laplacian &220.23 &49.01M & 73.1 (+0.4) \\
\multirow{7}{*}{(Ours)}
&Laplacian + $L$ &220.23 &49.01M & 73.6 (+0.9) \\
&DAConv  &220.23 &49.01M & 73.7 (+1.0)\\
&DAConv + $L$ & 220.23 &49.01M & 74.4 (+1.7) \\
&Conv$1\times 1$  &220.24 &48.98M &73.5 (+0.8) \\
&Conv$1\times 1$ + $L$ &220.24 &48.98M & 74.2 (+1.5) \\
&DAConv + PSP  & 220.55 &  50.36M & 74.0 (+1.3) \\
&DAConv + PSP + $L$ & 220.55 &  50.36M &  74.7 (+2.0)\\
\bottomrule[1.28pt]

\end{tabular}
}
\label{tab:saliencygeneration}
\vspace{-3.98mm}
\end{table}

\begin{table}[t!]
\centering
\caption{
Ablation study for the frequency modulation $L_{\text{FM}}$ and semantic high-frequency loss $L_{\text{SHF}}$.
}
\vspace{-3.98mm}
\scalebox{0.958}{
\begin{tabular}{c|ccccc}
\toprule[1.28pt]
$L_{\text{FM}}$ & $\times$  & $\checkmark$ & $\times$ & $\checkmark$ \\
$L_{\text{SHF}}$ & $\times$  & $\times$ & $\checkmark$ & $\checkmark$  \\
\midrule
mIoU & 74.0 & 74.4 & 74.5 &  \bf 74.7\\
\bottomrule[1.28pt]
\end{tabular}
}
\label{tab:lossablation}
\vspace{-3.98mm}
\end{table}

\begin{table}[t!]
\centering
\caption{
Ablation study for the loss weights $\lambda_{\text{FM}}$ and $\lambda_{\text{SHF}}$ of frequency modulation $L_{\text{FM}}$ and semantic high-frequency loss $L_{\text{SHF}}$.
}
\vspace{-3.98mm}
\scalebox{0.958}{
\begin{tabular}{c|ccccc}
\toprule[1.28pt]
$L_{\text{FM}}$ & $\lambda_{\text{FM}}=0.01$  & $\lambda_{\text{FM}}=0.1$ & $\lambda_{\text{FM}}=0.5$ & $\lambda_{\text{FM}}=1$ \\
\midrule
mIoU & 74.6 &\bf 74.7 & 74.5 & 74.4\\
\midrule
\midrule
$L_{\text{SHF}}$ & $\lambda_{\text{SHF}}=25$  & $\lambda_{\text{SHF}}=50$ & $\lambda_{\text{SHF}}=100$ & $\lambda_{\text{SHF}}=200$ \\
\midrule
mIoU & 74.5 & 74.6 &\bf 74.7 & 74.5\\
\bottomrule[1.28pt]
\end{tabular}
}
\label{tab:lossweightablation}
\vspace{-3.98mm}
\end{table}

\begin{table}[tb!]
\centering
\caption{Ablation study on the number of Local Pixel Relation Module (LPRM) in MSAU.
The $d=1$ indicates the dilation of LPRM is 1.
}
\vspace{-3.98mm}
\scalebox{0.8}{
\begin{tabular}{cccccccc|cc}
\toprule[1.28pt]
$d=1$ & $d=2$ & $d=4$ & $d=8$ & $d=16$ & $d=32$ & $d=64$ & $d=128$ &mIoU \\
\midrule
 & & & & &  & & & 74.7\\
\checkmark & & & & &  & & & 75.0\\
\checkmark & \checkmark & & & &  & & & 75.2 \\
\checkmark & \checkmark &\checkmark & & & &  & &75.4 \\
\checkmark & \checkmark &\checkmark &\checkmark & & & & & 75.5 \\
\checkmark & \checkmark &\checkmark &\checkmark &\checkmark & & & \ &75.7 \\
\checkmark & \checkmark &\checkmark &\checkmark &\checkmark &\checkmark & & & 75.8\\
\checkmark & \checkmark &\checkmark &\checkmark &\checkmark &\checkmark &\checkmark &  &\bf 76.0\\
\checkmark & \checkmark &\checkmark &\checkmark &\checkmark &\checkmark &\checkmark & \checkmark & 75.8\\
\bottomrule[1.28pt]
\end{tabular}
}
\label{tab:MSLPRM_dilation}
\vspace{-3.98mm}
\end{table}

{
\subsection{In Context with Related Works}
\label{sec:discusion_related}
In this section, we further discuss the intrinsic differences between the proposed method and related works in detail.

\vspace{+0.518mm}
\noindent{\bf Adaptive image downsampler.}
While our proposed method, Spatial Frequency Modulation (SFM), may appear similar to existing adaptive downsamplers such as Efficient Segmentation (ES)~\cite{2019efficientseg} and its improved versions LDS~\cite{2022learningtodown} and LZU~\cite{2023lzu}, there are fundamental differences that set SFM apart:
1) General applicability. ES is specifically designed for image downsampling and is applied to efficient low-resolution semantic segmentation. Therefore, it cannot be applied for full-resolution semantic segmentation. On the contrary, the proposed SFM can generally improve semantic segmentation and is applicable to any resolution, including low-resolution scenarios, making it versatile and suitable for various scenarios.
2) Optimization beyond object boundaries. While ES primarily focuses on improving object boundaries in downsampled images, SFM goes a step further. It aims to enhance both the high-frequency area and object boundaries in downsampled features. This is achieved through modulating high-frequency features to lower frequencies before downsampling, followed by demodulation back to high frequency. By effectively mitigating aliasing effects, SFM preserves more frequency information, resulting in finer segmentation results.
3) Solid motivation and theoretical basis. Unlike ES, which is driven by empirical hypotheses, our SFM is grounded in the fundamental principles of signal processing, particularly the Shannon-Nyquist sampling theorem~\cite{shannon1949communication, nyquist1928}. This theorem forms the basis of SFM's motivation, addressing aliasing degradation phenomena in segmentation models.

To illustrate the advantages of SFM, we provide a comparison with ES~\cite{2019efficientseg} and its improved versions LDS~\cite{2022learningtodown} and LZU~\cite{2023lzu} for efficient low-resolution semantic segmentation on the Cityscapes dataset. 
As shown in Table~\ref{tab:low_resolution}, although the proposed method is not specifically designed for efficient low-resolution semantic segmentation, it outperforms ES by 1.2  mIoU (55.2 \emph{vs.} 54.0) with much fewer GFLOPs (15.3 \emph{vs.} 44.2). Additionally, it demonstrates a 0.2 mIoU improvement compared with LDS~\cite{2022learningtodown} and LZU~\cite{2023lzu} with much fewer GFLOPs (15.3 \emph{vs.} 50.0, 45.0).

\vspace{+0.518mm}
\noindent{\bf Difference from CondConv and DySample.}
Neither CondConv~\cite{2019condconv} nor DySample~\cite{2023dysample} can address the specific issue of aliasing degradation in segmentation models. 
In contrast, the proposed SFM is designed to tackle this challenge. 
As shown in Table~\ref{tab:condconv_dysample}, CondConv and DySample~\cite{2023dysample} cannot reduce the aliasing ratio but even slightly increase it (from 42.7\% to 43.8\% and 42.8\%), while our SFM largely reduces the aliasing ratio from 42.7\% to 34.2\%, indicating that the proposed SFM effectively reduces the proportion of high-frequency signals that lead to aliasing degradation.
As a result, our SFM outperforms CondConv and DySample by a large margin of 2.2 and 2.5 mIoU, respectively.

CondConv primarily focuses on improving convolutional blocks, while SFM targets the enhancement of the downsampling operation between convolutional blocks, which is orthogonal and complementary to each other. As demonstrated in Table~\ref{tab:condconv_dysample}, SFM further improves CondConv from 73.8 to 76.8 mIoU.

Regarding DySample~\cite{2023dysample}, it operates on uniformly sampled features and cannot apply to non-uniformly sampled features output by ARS. 
MSAU in the SFM operates on non-uniformly sampled modulated features and demodulates them to recover high frequencies through non-uniform upsampling and modeling multi-scale pixel relations. 
Thus, combining ARS and DySample leads to a decrease of 0.2 mIoU compared to the ResNet-50 baseline, while combining ARS and MSAU effectively increases the mIoU by 3.3 compared to the ResNet-50 baseline, as shown in Table~\ref{tab:correlation}.

\vspace{+0.518mm}
\noindent{\bf Sampling-based methods.}
Compared with existing sampling-based works such as DUpsampling~\cite{dupsample}, PointRend~\cite{2020pointrend}, and SegFix~\cite{2020segfix}, the proposed SFM presents intrinsic differences. While PointRend~\cite{2020pointrend}, SegFix~\cite{2020segfix}, and previous methods like DenseCRF~\cite{2011densecrf} and GUM~\cite{2018gun} act as post-refinement methods for segmentation models, DUpsampling aims to enhance the upsampling process of segmentation models. In contrast, our proposed SFM targets alleviating aliasing degradation and enhancing both the downsampling and upsampling stages of segmentation models.
As indicated in Table~\ref{tab:segrefine}, SFM outperforms DenseCRF~\cite{2011densecrf}, GUM~\cite{2018gun}, DUpsampling~\cite{dupsample}, PointRend~\cite{2020pointrend}, and SegFix~\cite{2020segfix} by a margin of 0.5 to 1.3 mIoU. Moreover, the proposed SFM can easily be combined with these methods to improve them. Integrating SFM with SegFix~\cite{2020segfix} yields a further improvement from 80.5 to 81.2 mIoU.
}

{
\subsection{Ablation Study}
\label{sec:ablation}

In this section, we conduct ablation study experiments for Adaptive ReSampling (ARS) and Multi-Scale Adaptive Upsampling (MSAU), on Cityscapes~\cite{cityscapes2016}.
We use PSPNet~\cite{pspnet} due to its simplicity and efficiency, and ResNet-50~\cite{resnet2016} with a downsampling stride of 32$\times$ serves as the backbone.

\vspace{+0.518mm}
\noindent{\bf Integration of ARS and MSAU.}
ARS and MSAU {cannot function independently and depend on each other} to effectively implement the SFM framework.
ARS can cause feature to be misaligned with the input image due to non-uniform sampling, leading to inaccuracies in model optimization and potential classification errors~\cite{2021fapn}. This necessitates MSAU to reverse the modulation, \ie, demodulating the modulated low-frequency features back to high-frequency features and aligning the features with the input image.
In Table~\ref{tab:correlation}, using only ARS leads to a decrease of -0.5 mIoU, while using only MSAU results in an increase of +1.3 mIoU. However, when combined, they yield a substantial improvement of +3.3 mIoU.
This indicates that they depend on each other to fully exploit the effectiveness of the SFM framework.

\vspace{+0.518mm}
\noindent{\bf Number of ARS.}
Modern models usually employ a series of downsampling operations to gradually reduce feature map dimensions~\cite{resnet2016, 2021swin, 2022convnet}. Thus, we can use multiple ARS blocks before each downsampling operation. Specifically, in the backbone, we employ ARS before downsampling operations at $\text{stage1}\rightarrow\text{stage2}$, $\text{stage2}\rightarrow \text{stage3}$, and $\text{stage3}\rightarrow \text{stage4}$. 
The downsampling strides of $\text{stage1}$, 2, 3, 4 are 4, 8, 16, 32$\times$ respectively.
For feature maps before $\text{stage1}$, they are usually too shallow (\eg, Swin~\cite{2021swin} and ConvNeXt usually have only one patchfy layer for the first 4$\times$ downsampling on input images), hence we do not add an ARS before it.

As shown in Table~\ref{tab:ars_number}, each Adaptive ReSampling (ARS) block is capable of relieving aliasing and enhancing the model's adaptability, leading to performance improvement with an increasing number of ARS blocks. Finally, the three ARS blocks improve the accuracy by approximately +2.0 mIoU.

\vspace{+0.518mm}
\noindent{\bf Attention map generation.} 
Here, we discuss different methods of generating an attention map. 
Intuitively, we aim to assign higher attention values to high-frequency signals in the feature map. To achieve this, we combine the high-pass Laplacian filter with a convolution layer to serve as an attention generator, extracting high-frequency components in the feature map. This approach results in +0.4 mIoU, as shown in Table~\ref{tab:saliencygeneration}. 
However, it is still suboptimal due to the fixed kernel of the Laplacian filter.

To address this limitation, we propose the difference-aware convolution (DAConv) as an alternative. 
The learnable DAConv captures high-frequency signals through a central-to-surrounding operation, leading to +1.0 mIoU. 
By enlarging the perceptual field with the pyramid spatial pooling (PSP) module, combining DAConv and PSP achieves +1.3/2.0 mIoU with and without the extra supervision provided by the frequency modulation loss and semantic high-frequency loss, respectively.
Additionally, we explore using a $1\times 1$ convolution for attention map generation. Despite its simplicity, it results in an improvement of 0.8 points. 
This demonstrates that our adaptive resampler can work effectively without requiring careful design, thereby proving the effectiveness of our spatial frequency modulation strategy. 

\vspace{+0.518mm}
\noindent{\bf Frequency modulation and semantic high-frequency loss.} 
In Table \ref{tab:lossablation}, we evaluate the performance with and without frequency modulation ($L_{\text{FM}}$) and semantic high-frequency loss ($L_{\text{SHF}}$). We observe that incorporating $L_{\text{FM}}$ or $L_{\text{SHF}}$ results in a +0.4 or +0.5 mIoU improvement, respectively. Incorporating both $L_{\text{FM}}$ and $L_{\text{SHF}}$ results in the highest mIoU of 74.7, indicating that both components positively contribute to the segmentation performance.

In Table \ref{tab:lossweightablation}, we further investigated the influence of the loss weights ($\lambda_{\text{FM}}$ and $\lambda_{\text{SHF}}$) on the mIoU. For $L_{\text{FM}}$, we varied $\lambda_{\text{FM}}$ from 0.01 to 1, and for $L_{\text{SHF}}$, we varied $\lambda_{\text{SHF}}$ from 25 to 200. The results show that the choice of loss weights has a slight impact on the segmentation performance. Specifically, for $L_{\text{FM}}$, the best mIoU of 74.7 is achieved when $\lambda_{\text{FM}}$ is set to 0.1. Similarly, for $L_{\text{SHF}}$, the highest mIoU of 74.7 is obtained when $\lambda_{\text{SHF}}$ is set to 100.
These findings underscore the importance of both frequency modulation and semantic high-frequency loss in enhancing segmentation performance. Furthermore, the results show only minor fluctuations when adjusting the loss weights.

\vspace{+0.518mm}
\noindent{\bf Number of LRPM in MSAU.} 
Here, we investigate the number of Local Pixel Relation Modules (LPRM) in Multi-Scale Adaptive Upsampling (MSAU). The receptive field, critical for building local pixel relations, can be adjusted by the dilation of the LRPM. We use a series of LRPMs with different dilations, allowing MSAU to utilize local pixel relations at multiple scales to refine the upsampling process.
We note that the Non-Uniform Upsampling (NUU) in MSAU is used with Adaptive ReSampling (ARS) by default to reverse the coordinates deformation and keep the results aligned with the input.
As shown in Table~\ref{tab:MSLPRM_dilation}, gradually increasing the number of LPRMs as well as their dilations results in a steady improvement in segmentation accuracy.
When we use cascaded 7 LPRMs with dilations of (1, 2, 4, 8, 16, 32, 64) in MSAU, the accuracy increases to the best of 76.0.
These results confirm the effectiveness of the multi-scale relation for adaptive upsampling.



}

{\color{diff}
\section{Conclusion}

In this work, we quantitatively establish the relationship between aliasing and segmentation accuracy by using the ``aliasing ratio" metric. 
We observe a phenomenon referred to as ``\textit{aliasing degradation}," where segmentation performance degrades as a greater proportion of high-frequency signals are subjected to aliasing. 

Motivated by this, we have formulated a novel solution called the Spatial Frequency Modulation (SFM) framework. It comprises two key operations: modulation and demodulation. Modulation involves shifting high-frequency information to lower frequencies before downsampling, effectively alleviating aliasing. Meanwhile, demodulation restores the modulated features back to high frequencies during upsampling. This approach not only mitigates aliasing degradation but also preserves fine details and textures essential for accurate scene parsing.
We have developed the SFM framework with adaptive resampling (ARS) and multi-scale adaptive upsampling (MSAU), both lightweight and easily integrable modules. 
Our experimental results demonstrate the consistent improvement of various state-of-the-art semantic segmentation architectures.
Moreover, we extend the proposed method to various visual tasks including image classification, robustness against adversarial attacks, and instance/panoptic segmentation. 
Experimental results demonstrate its efficacy and broad applicability.



In the future, it is worthwhile to delve deeper into spatial frequency modulation for various computer vision applications, including object detection and tracking. 
Moreover, we consider further investigation into the characteristics of neural networks in the Fourier domain as a promising avenue for future research.
Our approach, grounded in addressing the fundamental issue of aliasing, provides valuable insights and a practical solution for improving dense prediction tasks. The SFM framework holds potential for broader applications and could stimulate further research into frequency-aware deep learning, ultimately advancing the capabilities of computer vision systems.
}

\section*{Acknowledgements}
This work was supported by the National Natural Science Foundation of China (62331006, 62171038, and 62088101), the Fundamental Research Funds for the Central Universities, and the JST Moonshot R\&D Grant Number JPMJMS2011, Japan.

\ifCLASSOPTIONcaptionsoff
  \newpage
\fi

{
\bibliographystyle{IEEEtran}
\bibliography{egbib}
}

\begin{IEEEbiography}[{\includegraphics[width=1in,height=1.25in,clip,keepaspectratio]{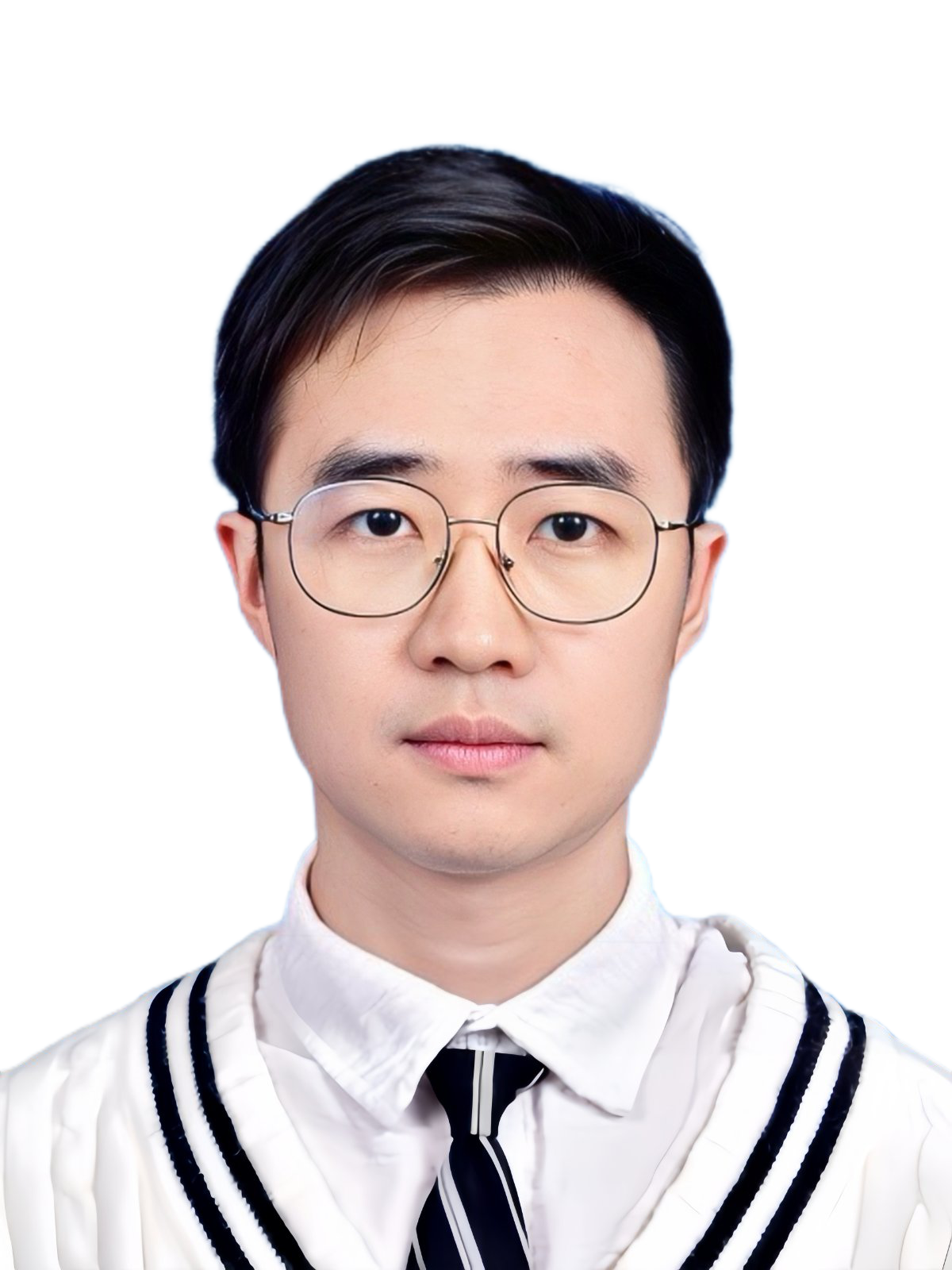}}]{Linwei Chen}
received the B.S. degree in mechanical engineering and automation from the China University of Geosciences, Beijing, China, in 2019, and the M.S. degree in software engineering from the Beijing Institute of Technology, Beijing, China, in 2021.
He is currently Eng.D. at MIIT Key Laboratory of Complex-field Intelligent Sensing, the School of Information and Electronics, Beijing Institute of Technology.
His research interests include image segmentation, object detection, and remote sensing.
\end{IEEEbiography}
\vspace{-3.98mm}
\begin{IEEEbiography}[{\includegraphics[width=1in,height=1.25in,clip,keepaspectratio]{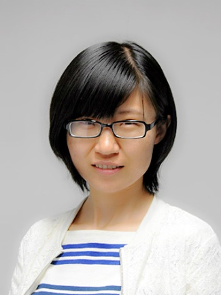}}]{Ying Fu}
received the B.S. degree in Electronic Engineering from Xidian University in 2009, the M.S. degree in Automation from Tsinghua University in 2012, and the Ph.D. degree in information science and technology from the University of Tokyo in 2015. 
She is currently a professor at the School of Computer Science and Technology, Beijing Institute of Technology. Her research interests include physics-based vision, image and video processing, and computational photography. 
\end{IEEEbiography}
\vspace{-3.98mm}
\begin{IEEEbiography}[{\includegraphics[width=1in,height=1.25in,clip,keepaspectratio]{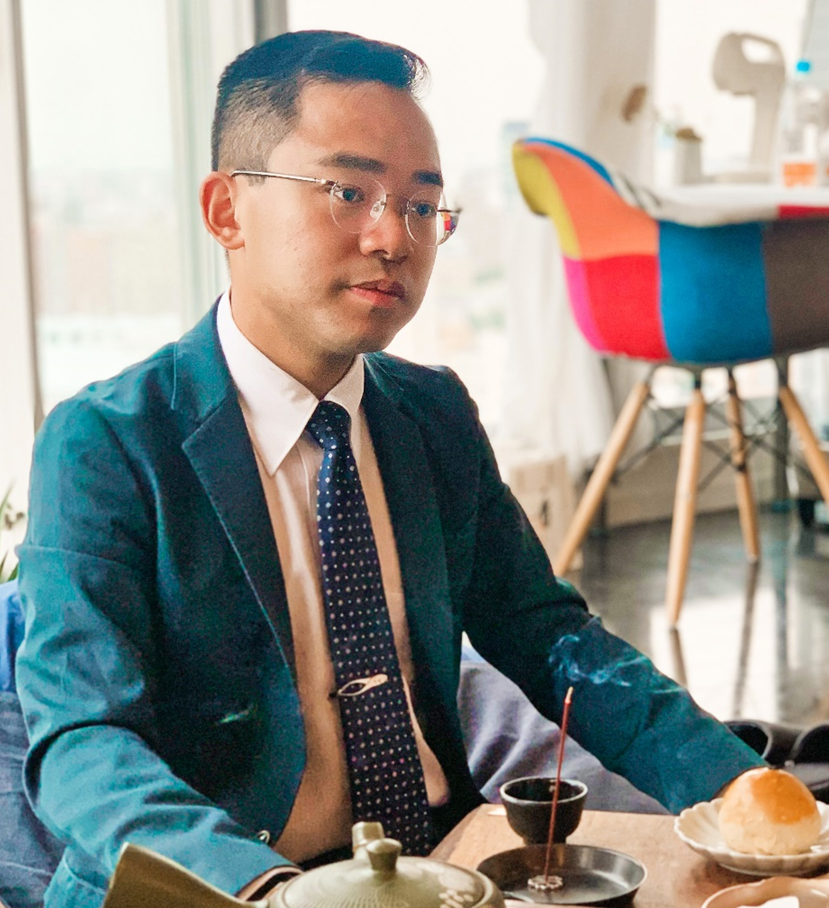}}]{Lin Gu}
completed his Ph.D. studies at the Australian National University and NICTA (Now Data61) in 2014.  After that, he was associated with the National Institute of Informatics in Tokyo and the Bioinformatics Institute, A*STAR, Singapore. Currently, he is now a research scientist at RIKEN AIP, Japan, and a special researcher at the University of Tokyo. He is also a project manager for Moonshot R\&D and the RIKEN-MOST program. His research covers a wide range of topics, encompassing computer vision, medical imaging, and AI for science.
\end{IEEEbiography}
\vspace{-3.98mm}
\begin{IEEEbiography}[{\includegraphics[width=1in,height=1.25in,clip,keepaspectratio]{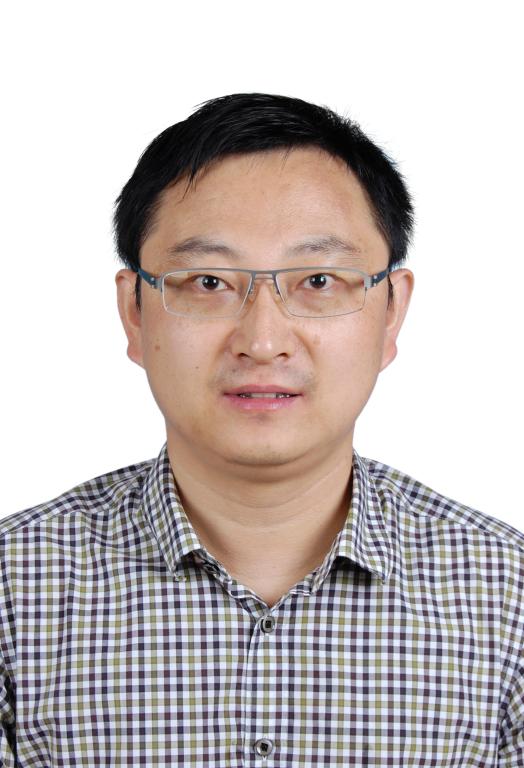}}]{Dezhi Zheng}
received the B.S. degree in mechanical engineering and the Ph.D. degree in precision instruments and mechanical engineering from Beihang University, Beijing, China, in 2000 and 2006, respectively. From 2014 to 2015, he was a Visiting Scholar at the University of Victoria, British Columbia, Canada. 
Since 2017, he has been a Professor with the Instrumentation and Optoelectronic Engineering, Beihang University. 
Since 2020, he has worked as a researcher at the MIIT Key Laboratory of Complex-Field Intelligent Sensing, Beijing Institute of Technology, Beijing, China.
He is the author of more than 100 articles and more than 40 inventions. 
His research interests include sensor technology and signal detection and processing technology.
\end{IEEEbiography}

\begin{IEEEbiography}[{\includegraphics[width=1in,height=1.25in,clip,keepaspectratio]{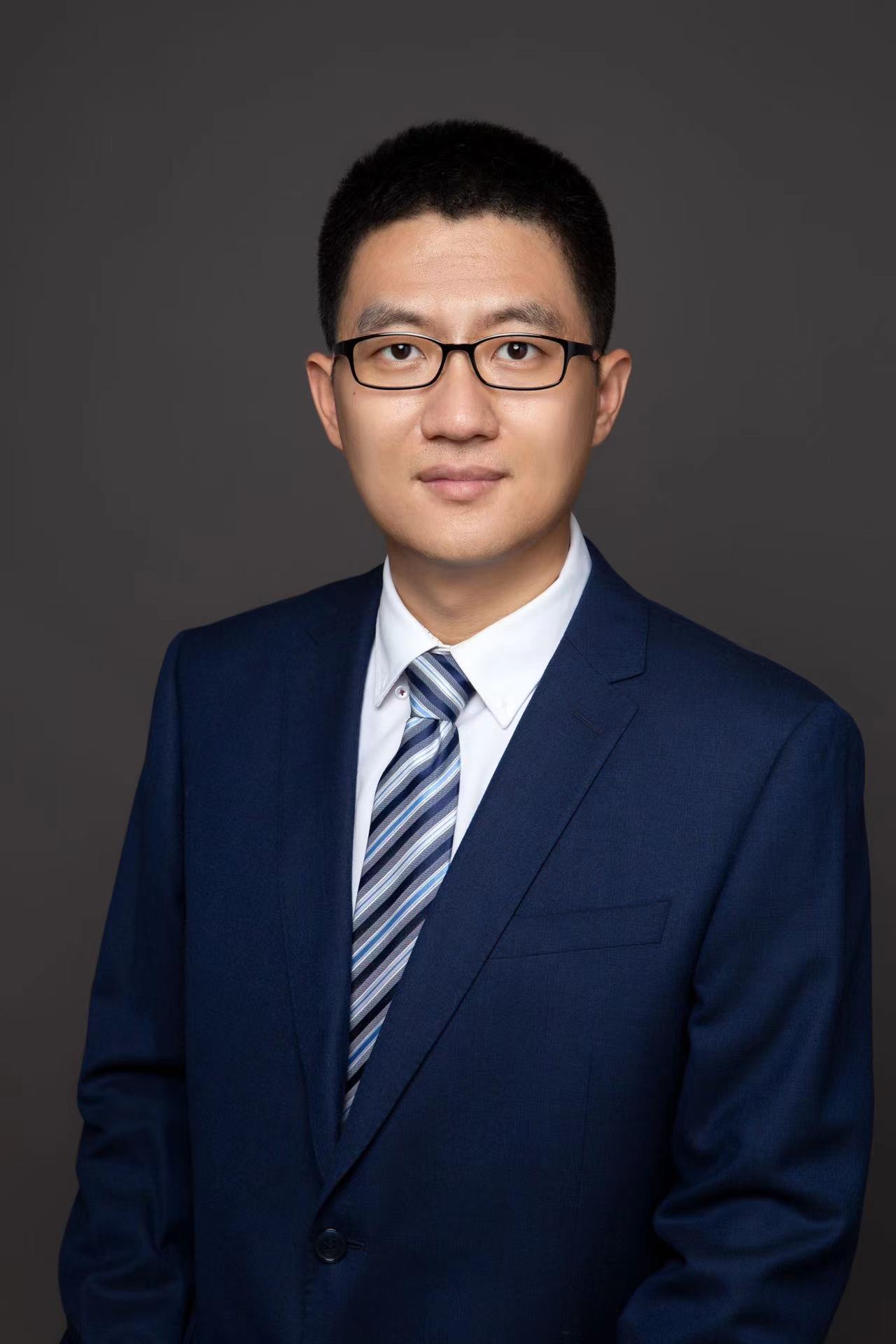}}]{Jifeng Dai}
is currently serving as an Associate Professor at the Department of Electronic Engineering at Tsinghua University. His primary research focus lies in the realm of deep learning for high-level vision.
Before his current position, he held the role of Executive Research Director at SenseTime Research, under the leadership of Professor Xiaogang Wang, from 2019 to 2022. Prior to that, he served as a Principal Research Manager in the Visual Computing Group at Microsoft Research Asia (MSRA) from 2014 to 2019, under the guidance of Dr. Jian Sun.
He earned his Ph.D. degree from the Department of Automation at Tsinghua University in 2014, under the supervision of Professor Jie Zhou. 
\end{IEEEbiography}

\end{document}